\documentclass{article}
\PassOptionsToPackage{table}{xcolor}
\usepackage{iclr2027_conference,times}

\usepackage{amsmath,amsfonts,bm}

\def\eqref#1{Eq.~\ref{#1}}
\def\1{\bm{1}}

\DeclareMathAlphabet{\mathsfit}{\encodingdefault}{\sfdefault}{m}{sl}
\SetMathAlphabet{\mathsfit}{bold}{\encodingdefault}{\sfdefault}{bx}{n}

\usepackage{graphicx}
\usepackage{xcolor}
\usepackage{booktabs}
\usepackage{tabularx}
\usepackage{wrapfig}
\usepackage{placeins}
\usepackage{needspace}
\usepackage{hyperref}
\usepackage{url}
\usepackage{caption}
\graphicspath{{figures/}}
\title{Know, then Draw: A New Learning Paradigm for Visual Generation}
\author{%
  \textbf{Yucheng Mao\textsuperscript{1,2}, Chongjian Ge\textsuperscript{2}, Zhuowen Tu\textsuperscript{1}, Hao Tan\textsuperscript{2}, Hanwen Jiang\textsuperscript{2}}\\[0.6ex]
  \normalfont\textsuperscript{1}University of California, San Diego\quad
  \textsuperscript{2}Adobe Research
}

\iclrfinalcopy

\begin{document}
\maketitle
\lhead{Preprint.} % Override the conference-publication status for arXiv.

\begin{center}
  \vspace{-0.05in}
  \includegraphics[width=\linewidth]{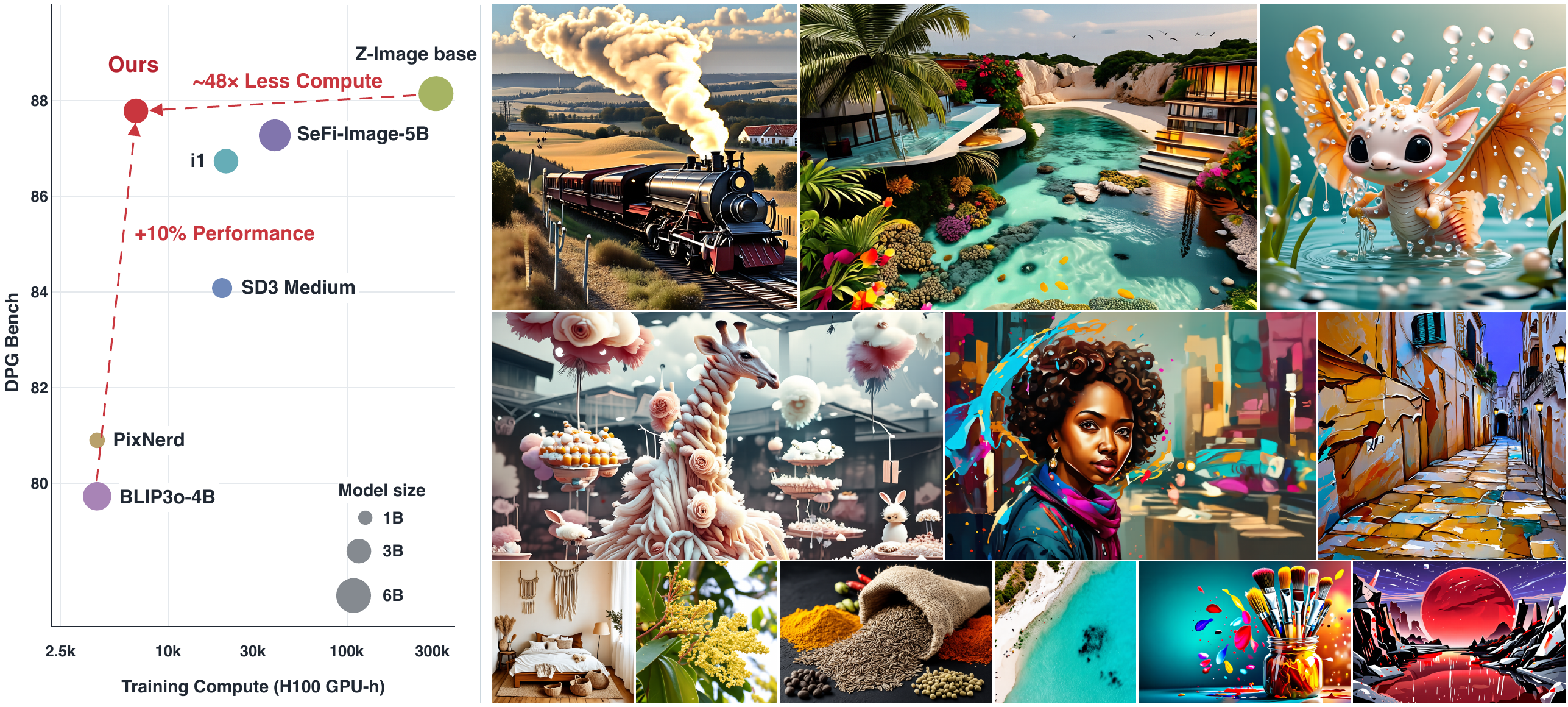}
  \vspace{-0.26in}
  \captionof{figure}{\small{
  \textbf{Strong visual generation with compute-efficient training.}
  \textbf{Left}: DPG-Bench versus training compute; marker size denotes model
  size. Our 3B model achieves competitive performance with only 6,600 H100
  GPU-hours (\textbf{$48\times$ less}  than Z-Image-Base). \textbf{Right}: Text-to-Image generated samples.
  }}
  \label{fig:teaser}
  % \vspace{-0.20in}
\end{center}

\FloatBarrier

\begin{abstract}
Visual generation requires both high-level semantic knowledge and fine-grained
pixel details. We propose a learning paradigm that assigns these roles to
autoregressive pretraining for knowledge acquisition and diffusion post-training
for high-fidelity rendering. 
Within this framework, we study the objective, representation, and scaling of knowledge acquisition and rendering.
Controlled comparisons identify autoregression as the stronger pretraining objective for downstream rendering, improving both learning efficiency and final generation
quality. We further find that  AR-only generation quality does not necessarily correlate with final performance; semantic quality matters more. This motivates a
purely semantic tokenizer trained with language supervision alone, without an
image decoder or reconstruction loss. We establish compute-optimal scaling laws
for knowledge acquisition and rendering, and find that autoregressive pretraining
lowers the fitted asymptotic rendering loss compared with pixel diffusion models trained from scratch. Guided by the laws, we train a 3B-parameter text-to-image model with only 6,600 H100 GPU-hours, achieving 86 on GenEval and 87 on DPG-Bench and validating the proposed  paradigm at scale.

% \keywords{Text-to-Image Generation \and Scaling Law \and Representation Learning}
% Together, our results support a simple division of labor: autoregression for
% knowledge acquisition and diffusion for rendering.
\end{abstract}

\Needspace{8\baselineskip}
\vspace{-2mm}
\section{Introduction}
\label{sec:intro}
\vspace{-1mm}

Visual generation has advanced primarily through two modeling paradigms:
autoregressive models predict discrete visual tokens~\citep{sun2024llamagen,tian2024var,pang2024randar}, while latent diffusion
models (LDMs) and pixel diffusion models (PDMs) denoise continuous signals~\citep{ho2020ddpm,rombach2022latentdiffusion,peebles2023dit}. They are
typically compared as competing paradigms, asking which produces better images.
Yet this comparison asks each paradigm to solve two distinct learning problems:
\textit{acquiring high-level visual knowledge} and \textit{rendering it into detailed images}.
We therefore argue that these problems need not favor the same objective, and look beyond
their competition to uncover the synergy between autoregression and diffusion.

% A longstanding
% debate concerns which paradigm provides a stronger foundation for visual
% generation. Yet this debate implicitly treats the two as end-to-end
% alternatives, asking each objective to solve the entire generation
% problem---from acquiring high-level visual knowledge to rendering low-level
% appearance.

% We propose to organize autoregression and diffusion into a unified
% \textbf{pretraining--post-training paradigm}.
% Their different prediction objectives may favor different stages
% of learning, allowing capabilities acquired under one objective
% to support those learned under the other.
% The central question is therefore how to assign knowledge acquisition
% and rendering to these objectives, and how to transfer what is learned
% between stages. This perspective turns the competition between
% autoregression and diffusion into a study of their complementary
% roles within a shared training process.

% We argue that visual knowledge
% acquisition and rendering are distinct learning problems and need not favor the
% same objective. We therefore ask a different question: rather than which
% paradigm is better for visual generation as a whole, which objective is best
% suited to each role when the two are optimized separately? This separation
% parallels the \textbf{pretraining--post-training paradigm} of modern language
% models, in which models first acquire broad knowledge and then adapt to desired
% behaviors. Their competition as generators may conceal their complementarity
% as learning objectives.

We find that the two problems indeed favor different objectives:
\textbf{autoregression for visual knowledge acquisition and diffusion for
rendering}. This specialization gives rise to a natural
\textbf{pretraining--post-training paradigm}, where autoregressive pretraining
provides the foundation on which diffusion is post-trained for rendering. This
mirrors modern language models, where broad knowledge is first acquired and
subsequently adapted toward desired behaviors. At inference, the autoregressive
rollout provides a structured visual representation that conditions diffusion,
connecting the two learning processes. Across the evaluated protocols and
compute budgets, this specialization consistently yields stronger generators,
revealing the synergy between the two objectives.

Specializing autoregression for knowledge acquisition changes what it should
learn. In visual autoregressive models, the tokenizer defines the
prediction target and therefore shapes the information the model is
incentivized to capture. Conventional tokenizers preserve low-level appearance
for image reconstruction. Once rendering is delegated to diffusion, however,
the autoregressive representation no longer needs to support reconstruction or
standalone image generation. Across different tokenizers, we find a striking
dissociation: neither reconstruction quality nor AR-only generation quality
determines post-training generation performance; semantic quality does. This
motivates \textbf{Semantic-LM}, a semantic tokenizer trained through language
supervision alone, without an image decoder or reconstruction loss. Visual
tokenization thus shifts from preserving pixels for reconstruction to providing
semantic learning targets for autoregressive knowledge acquisition.

Separating knowledge acquisition from rendering introduces a new question for scaling visual generation: how should compute be allocated between the two stages? We establish scaling laws for autoregressive pretraining and diffusion post-training that characterize the effects of model size, data, and compute and guide compute-optimal allocation. These laws also reveal a broader benefit of autoregressive pretraining. At matched model and data scales, diffusion post-training achieves both lower loss and a lower fitted asymptotic rendering loss than PDM trained from scratch. This finding suggests that autoregressive pretraining improves the attainable performance of diffusion rendering beyond accelerating convergence speed.

Given an available training budget of approximately 6,600 H100 GPU-hours, we
use our scaling laws to select a roughly 3B-parameter model and allocate compute
between pretraining and post-training. The resulting model achieves strong
text-to-image generation (Fig.~\ref{fig:teaser}), reaching 0.86 on
GenEval~\citep{ghosh2023geneval} and 87.78 on
DPG-Bench~\citep{hu2024ella}. This performance is achieved with approximately
\textbf{1/48} of the estimated training compute of
Z-Image-Base~\citep{zimage2025}, while attaining comparable DPG-Bench
performance.% These results validate the proposed learning paradigm at scale.

\vspace{-1mm}
\section{Method}
\label{sec:method}
\vspace{-1mm}

\begin{figure}[!t]
  \centering
  \setlength{\tabcolsep}{6pt}
  % \begin{tabular}{@{}cc@{}}
  %   \small Stage 1: AR pretraining & \small Stage 2: pixel-diffusion post-training \\
  %   \small (knowledge acquisition) & \small (rendering) \\[3pt]
  %   \draftpanel{1.15in}{0.46\linewidth}{%
  %     Text and discrete visual tokens enter a transformer that predicts the next
  %     visual token with a cross-entropy loss.\\[4pt]
  %     Architecture schematic: \resultTBD} &
  %   \draftpanel{1.15in}{0.46\linewidth}{%
  %     The AR backbone is frozen and its attention reused; added diffusion modules
  %     render pixels with a velocity-MSE loss.\\[4pt]
  %     Architecture schematic: \resultTBD}
  % \end{tabular}
  \includegraphics[width=\linewidth]{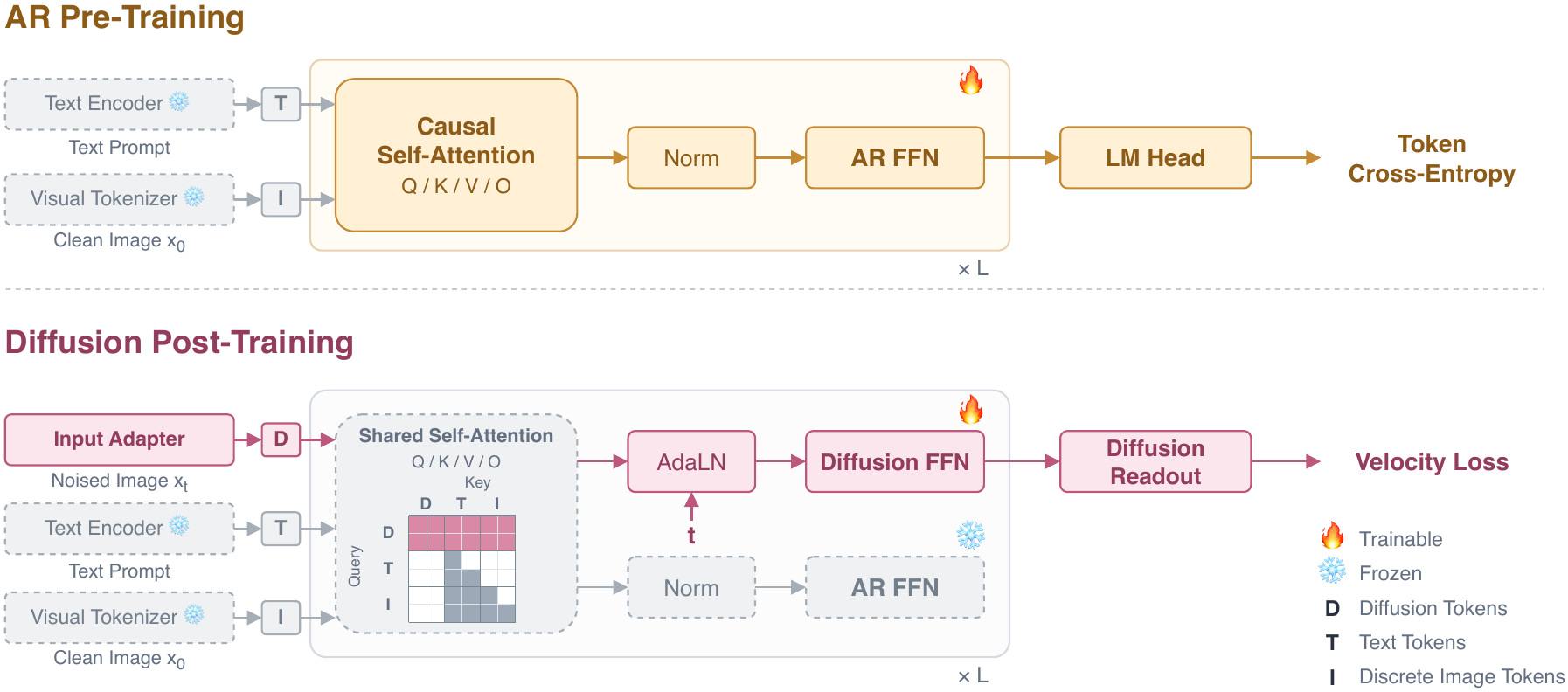}
  \vspace{-0.3in}
  \caption{\small{\textbf{Overview of our learning paradigm.} Autoregressive pretraining acquires visual knowledge through discrete token prediction. The pretrained backbone is then frozen and augmented with diffusion modules that reuse its attention and representations to render high-fidelity images. This division of labor specializes autoregression for knowledge acquisition and diffusion for rendering.}}
  \label{fig:method_overview}
  \vspace{-0.2in}
\end{figure}

As shown in Fig.~\ref{fig:method_overview}, our framework assigns visual
knowledge acquisition and rendering to distinct learning objectives. During
pretraining, a frozen visual tokenizer converts images into discrete tokens,
which a text-conditioned autoregressive (AR) model learns to predict. We then
freeze the AR backbone and add trainable pixel-diffusion modules that reuse its
attention projections and representations for rendering. During post-training,
the renderer conditions on tokens from ground-truth images; at inference, these
are replaced by tokens generated autoregressively from the text prompt.

% Our framework assigns knowledge acquisition and rendering to different
% objectives (Fig.~\ref{fig:method_overview}): a text-conditioned AR model learns discrete targets defined by a
% visual tokenizer, while pixel-diffusion post-training renders its learned
% representations into images. We freeze the pretrained AR backbone and augment it
% with diffusion-specific modules that reuse its attention and representations.
% During post-training, the renderer conditions on tokens from the ground-truth
% image; at inference, these are replaced by tokens autoregressively generated
% from the text prompt.

\vspace{-1mm}
\subsection{Autoregressive Pretraining for Knowledge Acquisition}
\vspace{-1mm}

\label{sec:method_ar}

\noindent \textbf{Autoregressive objective.}
Given an image and its caption, a frozen visual tokenizer maps the image to
discrete tokens $\mathbf{v}=(v_1,\ldots,v_L)$, while a frozen text encoder
produces conditioning features $\mathbf{c}$. Following
RandAR~\citep{pang2024randar}, we sample a permutation $\pi$ over spatial
positions and use position-instruction tokens $\mathbf{o}$ to specify each
prediction location. We train the AR model with
\begin{equation}
  \mathcal{L}_{\mathrm{AR}}(\theta)
  = -\sum_{i=1}^{L}\log p_\theta\!\left(v_{\pi(i)}\mid
      v_{\pi(<i)},\,\mathbf{o},\,\mathbf{c}\right),
  \label{eq:ar_ce}
\end{equation}
averaged over images and sampled permutations, where $v_{\pi(<i)}$ denotes
preceding tokens in the sampled order. Only the AR parameters $\theta$ are
optimized. By defining the prediction targets, the tokenizer determines what
information the model is trained to capture.

\begin{figure}[h]
  \centering
  % \vspace{-0.05in}
  \includegraphics[width=\linewidth]{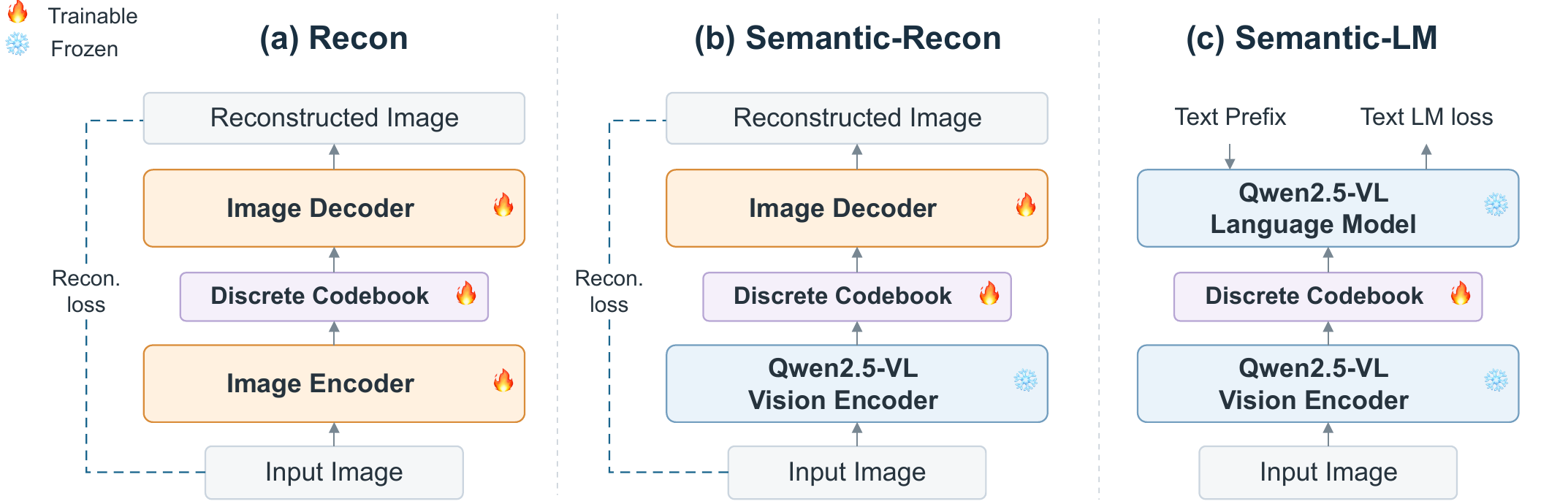}
  \vspace{-0.25in}
  \caption{\small{\textbf{Training the three visual tokenizers.}
  Recon and Semantic-Recon use image reconstruction supervision, whereas
  Semantic-LM learns its codebook through language modeling.}}
  \label{fig:tokenizer_training}
  \vspace{-0.05in}
\end{figure}

\noindent \textbf{Visual tokenizers.}
The visual tokenizer defines the prediction targets for AR pretraining and thus shapes the information the model learns. To examine which targets best support visual knowledge acquisition, we compare three tokenizers with different visual representations and training supervision (Fig.~\ref{fig:tokenizer_training}), including \textbf{Recon}, \textbf{Semantic-Recon}, and \textbf{Semantic-LM}.

\textbullet~~\textbf{Recon.}
We use the pretrained reconstruction-based Emu3.5 tokenizer~\citep{cui2025emu3}, which consists of an image encoder, an IBQ codebook~\citep{shi2025ibq}, and an image decoder.
\vspace{-0.05in}

\textbullet~~\textbf{Semantic-Recon.}
We quantize features of a frozen Qwen2.5-VL vision encoder~\citep{bai2025qwen25vltechnicalreport} using an IBQ codebook. We train the codebook and a decoder with reconstruction supervision.
\vspace{-0.05in}

\textbullet~~\textbf{Semantic-LM.}
We use the same semantic visual representation but train the IBQ codebook through next-token prediction within a frozen Qwen2.5-VL model. This variant requires neither an image decoder nor a reconstruction loss.

These variants let us examine two design choices. Comparing \textbf{Recon} with \textbf{Semantic-Recon} evaluates the benefit of using semantic representations for AR pretraining. Comparing \textbf{Semantic-Recon} with \textbf{Semantic-LM} tests whether reconstruction supervision is necessary for learning effective semantic tokens. All tokenizers remain frozen during generator training and are evaluated using the same protocol. We provide the tokenizer training details in Appendix~\ref{app:tokenizer_training}. 

\vspace{-2mm}
\subsection{Pixel-Diffusion Post-Training for Rendering}
\label{sec:method_diffusion}
\vspace{-2mm}

To render the representations learned during AR pretraining, we freeze the backbone parameters $\theta$ and introduce trainable pixel-diffusion modules $\phi$. The renderer reuses the backbone's frozen attention projections. Its trainable modules $\phi$ comprise an input projection for noisy image patches, timestep embeddings, diffusion-specific feed-forward and adaptive-normalization layers~\citep{peebles2023dit}, and a linear pixel readout. Only $\phi$ is optimized during post-training.

We train the renderer using pixel-space flow matching. Given a clean image $x_0$ and noise $\epsilon$, we construct $x_t=(1-t)x_0+t\epsilon$. The renderer receives $x_t$, timestep $t$, text features $\mathbf{c}$, and representations of visual tokens $\mathbf{v}$ produced by the frozen AR backbone.  The renderer predicts the clean image $\widehat{x}_0$, which we convert to a velocity prediction $f_\phi$ using a numerically stabilized transformation. We train the renderer by minimizing the mean squared error between the predicted and target velocities:
\begin{equation}
  \mathcal{L}_{\mathrm{Diff}}(\phi)
  = \mathbb{E}_{(x_0,\mathbf{v},\mathbf{c}),\epsilon,t}
    \left[\big\lVert f_\phi(x_t,t,\mathbf{v},\mathbf{c};\theta)
      -v_{\mathrm{target}}(x_t,x_0,t)\big\rVert_2^2\right].
  \label{eq:post_vmse}
\end{equation}
The loss is computed in velocity space, while the network directly predicts $\widehat{x}_0$. We sample timesteps from a logit-normal distribution with a resolution-dependent shift~\citep{esser2024scalingrectifiedflow}. Appendix~\ref{app:tokenizer_controls} details the conversion from image predictions to velocities and the optimization settings.

The source of the conditioning tokens differs between post-training and inference. During post-training, we extract tokens directly from the ground-truth image $x_0$, so the visual condition corresponds to the image used to construct $x_t$. At inference, the AR model generates visual tokens from the text prompt, and these tokens condition the renderer.

\subsection{Scaling Methodology}
\label{sec:method_scaling}
Separating knowledge acquisition from rendering introduces a resource allocation problem: given a fixed compute budget, how much should be assigned to each stage, and how should each stage balance model size and training data? We address these questions by fitting scaling laws for autoregressive pretraining and diffusion post-training. We also fit scaling law for a pixel-diffusion baseline trained from scratch to examine the benefits of autoregressive pretraining.

\noindent \textbf{Fitting objective-specific scaling laws.}
For each objective, we train a family of models across widths and depths and fit
\begin{equation}
  \widehat{\mathcal{L}}(N,D)
  = E
  + A\left(\frac{N}{N_0}\right)^{-\alpha}
  + B\left(\frac{D}{D_0}\right)^{-\beta},
  \label{eq:scaling_form}
\end{equation}
where $N$ is the number of trainable parameters, $D$ is the cumulative training-token count, and $E$ estimates the asymptotic loss under the fitted model. We fit autoregressive cross-entropy and diffusion velocity-MSE separately, using $C=6ND$ as a compute proxy. For diffusion post-training, this proxy excludes the frozen AR forward pass (Appendix~\ref{app:post_protocol}). We cross-check the parametric fits using training-curve envelopes and IsoFLOP analyses.

To reduce optimization-related confounds, we tune training recipes that transfer across model sizes within each family. We also conduct an independent hyperparameter search for the from-scratch diffusion baseline. Appendices~\ref{app:ar_transfer} and~\ref{app:pdm_hp} describe hyperparameter transfer. Appendices~\ref{app:ar_protocol} and~\ref{app:post_scaling_results} provide compute accounting and the full scaling protocols.

\noindent \textbf{Optimizing the learning paradigm through compute allocation.}
Within our two-stage framework, compute allocation determines the relative emphasis on knowledge acquisition and rendering. We parameterize this allocation by
$r=C_{\mathrm{post}}/(C_{\mathrm{pre}}+C_{\mathrm{post}})$, so that different values of $r$ define different learning paradigm. To derive a scaling-guided allocation, we first minimize each stage's fitted loss under the compute constraint $C=6ND$. For each paired AR backbone and diffusion renderer, we then determine the data budget required to place each stage at its compute-optimal operating point. The resulting compute costs specify the allocation $r$, which we examine across total budgets. This procedure translates the stage-specific scaling laws into a concrete training strategy without directly combining autoregressive and diffusion losses. We additionally compare the fitted asymptotic losses of diffusion post-training and pixel diffusion trained from scratch to assess potential benefits beyond faster convergence. Sec.~\ref{sec:exp_scaling} presents the results.

\vspace{-1mm}
\section{Experiments}
\label{sec:experiments}
\vspace{-2mm}

Our experiments address three questions. First, does autoregressive pretraining lead to better image generation than diffusion-based training under comparable model sizes and image exposure? (Sec.~\ref{sec:exp_sample_efficiency}) Second, should the visual tokens used for pretraining preserve image details or focus on semantic information? (Sec.~\ref{sec:exp_tokenizers}) Third, how should model size, training data, and compute be allocated between knowledge acquisition and rendering? (Sec.~\ref{sec:exp_scaling}) We then use these findings to guide the training of a larger text-to-image model and evaluate its generation quality (Sec.~\ref{sec:exp_hero}).

\vspace{-1mm}
\subsection{Autoregression or Diffusion for Visual Knowledge Acquisition?}
\label{sec:exp_sample_efficiency}
\vspace{-1mm}

We assess visual knowledge acquisition through the quality of image generation after pixel-diffusion post-training. We compare latent diffusion and autoregression as pretraining objectives, followed by the same rendering architecture and post-training protocol. We denote these two-stage routes as \emph{LDM$\xrightarrow{}$PDM} and \emph{AR$\xrightarrow{}$PDM}. We also include pixel diffusion and latent diffusion as baselines.

\noindent \textbf{Training and Evaluation Setup.}
All four routes use approximately 400M parameters in the final generator and the same cumulative number of training image samples.  Both two-stage routes combine a 250M-parameter pretrained backbone with 150M trainable rendering parameters. They also use semantic visual representations: AR uses our semantic tokenizer, while LDM pretraining uses SVG Autoencoder-P, a semantic tokenizer built on DINOv3~\citep{shi2025svgt2i,simeoni2025dinov3}. The former produces discrete tokens, whereas the latter produces continuous representations. The direct LDM baseline uses the reconstruction-based FLUX.2 autoencoder~\citep{bfl2025flux2}. We evaluate generation quality on GenEval~\citep{ghosh2023geneval} and DPG-Bench~\citep{hu2024ella}. Tab.~\ref{tab:experiment_recipe} and Appendix~\ref{app:tokenizer_controls} provide the full training and evaluation protocols.

% All routes use matched cumulative image exposure and approximately 400M
% parameters in the final generator. For LDM$\xrightarrow{}$PDM and AR$\xrightarrow{}$PDM, we further align
% semantic visual representations and model capacity: each uses an approximately
% 250M pretrained backbone with 150M pixel-rendering parameters. AR uses our
% semantic tokenizer, while LDM uses SVG Autoencoder-P, a semantic tokenizer built on
% DINOv3~\citep{shi2025svgt2i,simeoni2025dinov3}; the direct LDM baseline instead uses the
% reconstruction-based FLUX.2 autoencoder~\citep{bfl2025flux2}. SVG provides
% continuous semantic representations, whereas our semantic tokenizer produces
% discrete semantic tokens. We evaluate with
% GenEval~\citep{ghosh2023geneval} and DPG-Bench~\citep{hu2024ella}.
% Tab.~\ref{tab:experiment_recipe} and
% Appendix~\ref{app:tokenizer_controls}
% provide complete protocols.

\noindent \textbf{Results.}
Autoregressive pretraining yields better generation quality after  diffusion post-training than the evaluated LDM-pretraining pipeline. AR$\xrightarrow{}$PDM reaches 64.8 on GenEval after 100k post-training
updates, exceeding the score of PDM trained for 400k iterations (61.2). With further post-training, it reaches 65.5 on GenEval and 71.1 on DPG-Bench (Tab.~\ref{tab:efficiency_results}). In contrast, direct LDM reaches 57.3 on GenEval, while LDM$\xrightarrow{}$PDM achieves 56.6.
These results indicate that splitting training into two stages alone is insufficient for generation. Under the evaluated settings, autoregression provides a more effective pretraining objective for subsequent pixel-diffusion rendering.

% Figures commented out (kept for reference); the dissociation result is now the
% rFID column of Tab.~\ref{tab:tokenizer_results} in Sec 4.2, and the efficiency
% curve is dropped in favor of Tab.~\ref{tab:efficiency_results}.
% \input{figures/efficiency_dissociation}

\begin{table}[t]
  \centering
  \caption{\small{\textbf{Comparison between different learning paradigms under a matched data budget.}
  Final-generator scores as percentages ($\uparrow$); GenEval Overall is the
  mean of the six category scores.}}
  \vspace{-0.1in}
  \label{tab:efficiency_results}
  \small
  \setlength{\tabcolsep}{2pt}
  \renewcommand{\arraystretch}{1.2}
  \begin{tabular*}{\linewidth}{@{\extracolsep{\fill}}l*{8}{c}@{}}
    \toprule
     Method & \multicolumn{7}{c}{\textbf{GenEval} $\uparrow$} & \cellcolor{gray!10}\textbf{DPG} $\uparrow$ \\
    \cmidrule(lr){2-8}
     & \cellcolor{gray!10}Overall & Single Obj. & Two Obj. & Counting & Colors & Position & Color Attri. & \cellcolor{gray!10} \\
    \midrule
    PDM       & \cellcolor{gray!10}61.16 & 97.50 & 47.47 & 33.75 & 87.23 & 50.00 & 51.00 & \cellcolor{gray!10}62.56 \\
    LDM       & \cellcolor{gray!10}57.34 & 97.50 & 55.56 & 27.50 & 74.47 & 38.00 & 51.00 & \cellcolor{gray!10}64.33 \\
    LDM $\xrightarrow{}$ PDM & \cellcolor{gray!10}56.61 & 90.00 & 60.61 & 32.50 & 75.53 & 39.00 & 42.00 & \cellcolor{gray!10}60.01 \\
    AR $\xrightarrow{}$ PDM (ours)  & \cellcolor{gray!10}\textbf{65.51} & 97.50 & 66.67 & 40.00 & 81.91 & 51.00 & 56.00 & \cellcolor{gray!10}\textbf{71.11} \\
    \bottomrule
  \end{tabular*}
  \vspace{-0.14in}
\end{table}

\vspace{-1mm}
\subsection{What Should Autoregressive Pretraining Predict?}
\vspace{-1mm}
\label{sec:exp_tokenizers}

The preceding experiments support autoregression as an effective pretraining objective for subsequent diffusion rendering. We next ask what information the AR model should learn. Since the tokenizer defines its prediction targets, we compare three tokenizers: \textbf{Recon}, \textbf{Semantic-Recon}, and \textbf{Semantic-LM} (Sec.~\ref{sec:method_ar}). These variants allow us to examine whether effective targets need to preserve image details for reconstruction or can focus on semantic information. We keep generator training settings fixed and evaluate final image generation quality after the same pixel-diffusion post-training protocol. Appendix~\ref{app:tokenizer_training} provides the tokenizer training details.

% Tokenizer defines the autoregressive prediction target, we next
% study what information AR pretraining should learn. We compare
% \textbf{Recon}, \textbf{Semantic-Recon}, and \textbf{Semantic-LM}
% (Sec.~\ref{sec:method_ar}) under identical training
% settings, evaluating each representation by final generation quality after the
% same pixel-diffusion post-training. Tokenizer-training details are provided in
% Appendix~\ref{app:tokenizer_training}.

\begin{table}[t]
  \centering
  \caption{\small{\textbf{Semantic quality, rather than reconstruction or AR-only
generation quality, determines final generation performance.}
Recon performs best in reconstruction and AR-only generation, yet produces the
weakest final generator; Sem-LM remains strong downstream despite having no
reconstruction supervision or image decoder. $\Delta$ denotes the change after
diffusion post-training.}}
\vspace{-0.11in}
  \label{tab:tokenizer_results}
  \small
  \setlength{\tabcolsep}{6pt}
  \renewcommand{\arraystretch}{1.2}
  \begin{tabular}{@{}ll*{7}{c}@{}}
    \toprule
    \multicolumn{3}{c}{Tokenizer Properties} &
    \multicolumn{2}{c}{AR-only Generation} &
    \multicolumn{4}{c}{Post-train Diffusion Generation} \\
    \cmidrule(lr){1-3}
    \cmidrule(lr){4-5}
    \cmidrule(lr){6-9}
    Tokenizer & Supervision & rFID$\downarrow$ &
    GenEval$\uparrow$ & DPG$\uparrow$ &
    GenEval$\uparrow$ & $\Delta$ GenEval & DPG$\uparrow$ & $\Delta$ DPG \\
    \midrule
    Recon     & Recon & 1.04     & 57.49 & 65.94 & 54.67 &
    \textcolor{red!80!black}{$-2.82$} & 62.98 & \textcolor{red!80!black}{$-3.04$} \\
    Sem-Recon & Recon & 40.5     & 50.44 & 67.03 & 65.51 &
    \textcolor{green!55!black}{$+15.07$} & 71.11 & \textcolor{green!55!black}{$+4.08$} \\
    Sem-LM    & LM    & $\infty$ & 0.00  & 0.00 & 64.84 &
    \textcolor{green!55!black}{$+64.84$} & 67.72 & \textcolor{green!55!black}{$+67.72$} \\
    \bottomrule
  \end{tabular}
  \vspace{-0.13in}
\end{table}

\begin{figure}[t]
  \centering
  \includegraphics[width=0.99\linewidth]{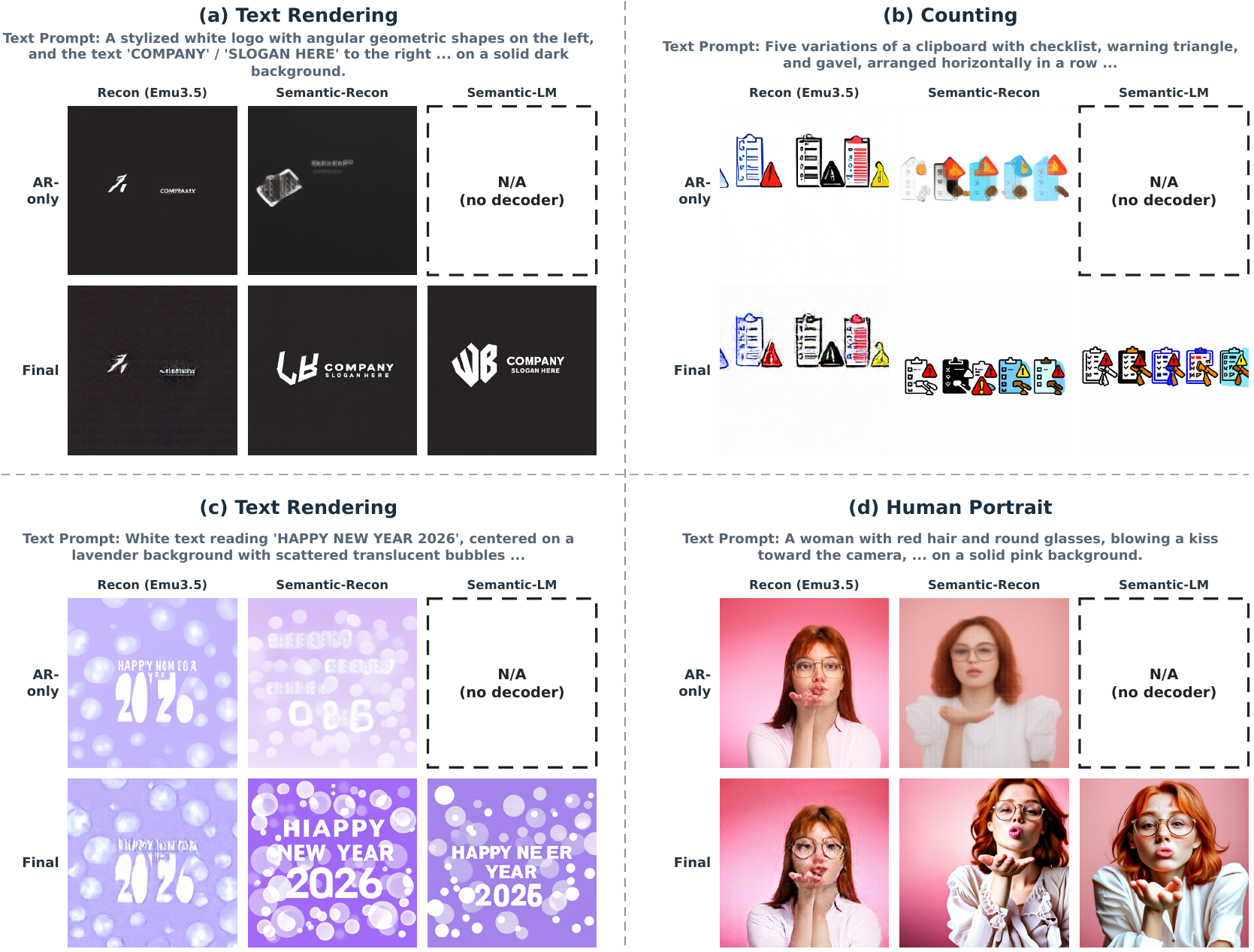}
  \vspace{-0.13in}
  \caption{\small{\textbf{AR-only generation quality does not predict final generation quality.}
    Top: AR-only decodes; bottom: final images after diffusion post-training.
    The semantic tokenizers produce stronger final images despite weaker or
    unavailable AR-only decodes. Semantic-LM has no  decoder.}}
  \label{fig:ar_stage_quality}
  \vspace{-0.425in}
\end{figure}

\noindent \textbf{Better reconstruction does not necessarily improve final generation.}
\textbf{Recon} achieves substantially better reconstruction quality than
\textbf{Semantic-Recon}, yet yields lower final generation scores:
54.7 versus 65.5 on GenEval and 63.0 versus 71.1 on DPG-Bench
(Tab.~\ref{tab:tokenizer_results}).
Moreover, \textbf{Semantic-LM} requires no reconstruction supervision,
yet achieves 64.8 on GenEval and 67.7 on DPG-Bench after post-training.
Results indicates effective AR targets need not support reconstruction.

\noindent \textbf{Better AR-only generation does not necessarily improve final generation.}
A similar pattern appears when comparing AR-only generation quality. \textbf{Recon} achieves a higher AR-only GenEval score than \textbf{Semantic-Recon} (57.49 versus 50.44), but this ranking reverses after diffusion post-training (54.67 versus 65.51), while DPG-Bench shows the same trend. Fig.~\ref{fig:ar_stage_quality} illustrates this contrast qualitatively. Although \textbf{Semantic-LM} has no image decoder and cannot generate images on its generated tokens, but still support effective diffusion rendering.

\noindent \textbf{Semantic representations provide effective pretraining targets.}
Together, these results show that neither reconstruction quality nor AR-only generation quality reliably predicts final generation quality across the evaluated tokenizers. The largest improvement comes from replacing reconstruction-oriented tokens with pretrained semantic representations. Removing reconstruction supervision retains much of this benefit. When diffusion handles rendering, AR pretraining can therefore focus on learning semantic information without requiring its prediction targets to support image generation.

\vspace{-2mm}
\subsection{Scaling Knowledge Acquisition and Rendering}
\label{sec:exp_scaling}
\vspace{-1mm}

The preceding experiments support autoregressive pretraining for knowledge acquisition and semantic tokens as effective prediction targets. We now ask how pretraining and rendering should scale as more compute becomes available. We fit scaling laws for both stages to study how model size and training data affect performance. We then compare diffusion post-training with training from scratch and use the fitted laws to guide compute allocation between the two stages.

\begin{figure}[t]
 \centering
 \includegraphics[width=\linewidth]{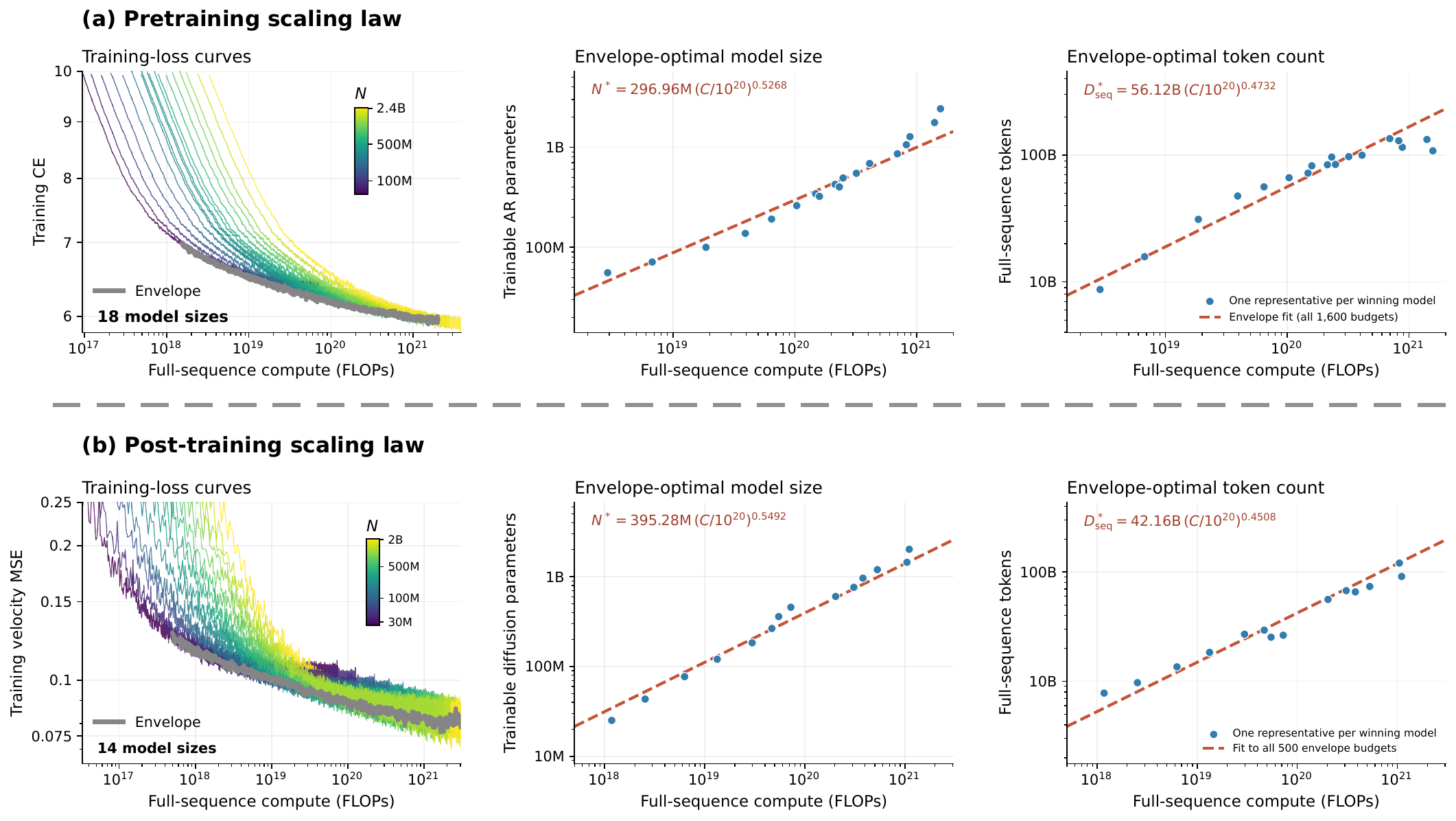}
 \vspace{-0.3in}
 \caption{\small{\textbf{Scaling laws for Pretraining and Post-training.}
 (a) AR pretraining; (b) pixel-diffusion post-training.
 Left: training loss versus full-sequence compute on logarithmic axes, with
 the lower envelope highlighted in gray; color bars indicate model size.
 Middle and right: envelope-optimal model and token allocations.
 Fitting protocols are given in Appendix~\ref{app:ar_protocol}
 and Appendix~\ref{app:post_scaling_results}.}}
 \label{fig:combined_scaling}
 \label{fig:ar_scaling}
 \label{fig:post_scaling}
 \vspace{-0.18in}
\end{figure}

\noindent \textbf{Different Scaling Behavior for Knowledge Acquisition and Rendering.}
We fit the autoregressive pretraining law using 18 models spanning
55.71M--2.43B trainable parameters (Fig.~\ref{fig:combined_scaling} (a)):
\begin{equation}
 \widehat{\mathcal L}_{\rm pre}(N,D_{\rm seq})
 =5.72+0.32\left(\frac{N}{10^8}\right)^{-0.47}
       +1.73\left(\frac{D_{\rm seq}}{10^9}\right)^{-0.44}.
 \label{eq:ar_scaling_candidate}
\end{equation}
The training-curve envelope, IsoFLOP, and parametric estimates yield
$N_{\mathrm{opt}}\propto C^{0.47\text{--}0.53}$ and
$D_{\mathrm{opt}}\propto C^{0.47\text{--}0.53}$.
These estimates suggest that model size and training data should grow at approximately the same rate as the pretraining budget increases.

We fit the post-training law to 14 models spanning 25.1M--2.03B trainable~parameters:
\begin{equation}
 \widehat{\mathcal L}_{\rm post}(N,D_{\rm seq})
 =0.070+0.022\left(\frac{N}{10^8}\right)^{-0.25}
       +0.013\left(\frac{D_{\rm seq}}{10^{10}}\right)^{-0.60}.
 \label{eq:post_scaling_candidate}
\end{equation}
The parametric fit favors increasing model size more rapidly than training data. Across the three estimation methods, however, the scaling exponents vary:
$N_{\mathrm{opt}}\propto C^{0.45\text{--}0.70}$ and
$D_{\mathrm{opt}}\propto C^{0.30\text{--}0.55}$.
Each renderer is paired with a correspondingly scaled frozen AR backbone, so this law describes rendering performance within the paired model family, rather than diffusion capacity alone.

These results suggest that pretraining and rendering can benefit from different balances between model size and training data. We therefore optimize the two stages separately before deriving their compute allocation. Appendix~\ref{app:ar_protocol} and Appendix~\ref{app:post_scaling_results} provide the full fitting protocols and results.

\noindent \textbf{Autoregressive Pretraining Lowers the Fitted Rendering Loss Floor.}
% \label{sec:exp_loss_floor}
We next examine whether the pretrained rendering pipeline offers benefits beyond faster convergence. We compare the fitted asymptotic losses of diffusion post-training and pixel diffusion trained from scratch under their respective conditioning protocols. Post-training yields $E_{\mathrm{post}}\approx0.070$ in velocity-MSE (Eq.~\ref{eq:post_scaling_candidate}). For independently tuned pixel diffusion trained from scratch, we obtain
\begin{equation}
 \widehat{\mathcal L}_{\mathrm{scratch}}(N,D_{\rm seq})
 =0.0788+0.0133\left(\frac{N}{10^8}\right)^{-0.3450}
       +0.0079\left(\frac{D_{\rm seq}}{10^{10}}\right)^{-0.8627},
 \label{eq:scratch_scaling}
\end{equation}
with $E_{\mathrm{scratch}}\approx0.0788$. This fit uses eight models with 23.83M--1.76B parameters. Appendix~\ref{app:pdm_scratch} reports the allocation estimates from all three methods.

The lower fitted asymptote suggests that the AR-then-diffusion pipeline's advantage may extend beyond faster convergence, with a lower estimated loss as model size and training data increase.  This comparison evaluates the pipeline as a whole, including both pretrained weights and conditioning tokens extracted from the target image, and does not isolate their individual contributions. The asymptotic term $E$ summarizes this extrapolated trend under the fitted scaling model.

\begin{table}[t]
  \centering
\vspace{-0.12in}

  \caption{\small{\textbf{Text-to-image generation on GenEval and DPG-Bench.}
  Model size is in billions of parameters (B), as reported for each system;
  component accounting may differ.  Dash denotes
 unavailable results. We use LLM to rewrite prompt in order to match the training distribution.}}
 \vspace{-0.1in}

  \label{tab:hero_results}
  \footnotesize
  \setlength{\tabcolsep}{2pt}
  \renewcommand{\arraystretch}{1.12}
  \begin{tabular*}{\linewidth}{@{\extracolsep{\fill}}l*{9}{c}@{}}
    \toprule
    & & \multicolumn{7}{c}{GenEval} & DPG \\
    \cmidrule(lr){3-9}\cmidrule(lr){10-10}
    Model & Size (B) & Single & Two & Count & Colors & Pos.
      & C-attr. & Overall$\uparrow$ & Overall$\uparrow$ \\
    \midrule
    PixArt-$\alpha$~\citep{chen2023pixart}   & 0.6 & 0.98 & 0.50 & 0.44 & 0.80 & 0.08 & 0.07 & 0.48 & 71.11 \\
    SDXL~\citep{podell2023sdxl}              & 2.6 & 0.98 & 0.74 & 0.39 & 0.85 & 0.15 & 0.23 & 0.55 & 74.65 \\
    DALL-E 3~\citep{betker2023dalle3}          & --  & 0.96 & 0.87 & 0.47 & 0.83 & 0.43 & 0.45 & 0.67 & 83.50 \\
    PixNerd-XXL/16~\citep{wang2026pixnerd}    & 1.2 & 0.97 & 0.86 & 0.44 & 0.83 & 0.71 & 0.53 & 0.73 & 80.90 \\
    SD3-Medium~\citep{esser2024scalingrectifiedflow}        & 2.0 & 0.99 & 0.94 & 0.72 & 0.89 & 0.33 & 0.60 & 0.74 & 84.08 \\
    BLIP3o-4B~\citep{chen2025blip3o}         & 4.0 & --   & --   & --   & --   & --   & --   & 0.77 & 79.73 \\
    FLUX.1-dev~\citep{bfl2024flux1dev}        & 12  & 0.98 & 0.93 & 0.75 & 0.93 & 0.68 & 0.65 & 0.82 & 84.00 \\
    HiDream-I1-Full~\citep{cai2025hidream}   & 17  & 1.00 & 0.98 & 0.79 & 0.91 & 0.60 & 0.72 & 0.83 & 85.89 \\
    Seedream 3.0~\citep{gao2025seedream3}      & --  & 0.99 & 0.96 & 0.91 & 0.93 & 0.47 & 0.80 & 0.84 & 88.27 \\
    i1~\citep{zeng2026i1}                & 3.0 & --   & --   & --   & --   & --   & --   & 0.84 & 86.73 \\
    Z-Image-Turbo~\citep{zimage2025}     & 6.0 & 1.00 & 0.95 & 0.77 & 0.89 & 0.65 & 0.68 & 0.82 & 84.86 \\
    Qwen-Image~\citep{wu2025qwenimage}        & 20  & 0.99 & 0.92 & 0.89 & 0.88 & 0.76 & 0.77 & 0.87 & 88.32 \\
    BLIP3o-NEXT SFT~\citep{chen2025blip3onext}   & 3.0 & --   & --   & --   & --   & --   & --   & 0.87 & 81.98 \\
    SeFi-Image-5B~\citep{sefi2026image}     & 5.0 & 1.00 & 0.92 & 0.87 & 0.90 & 0.84 & 0.76 & 0.88 & 87.27 \\
    \midrule
    \textbf{Ours}     & $\sim$3 & 1.00 & 0.82 & 0.75 & 0.99 & 0.79 & 0.81 & 0.86 & 87.78 \\
    \bottomrule
  \end{tabular*}
  \vspace{-0.07in}
\end{table}

\vspace{-1mm}
\noindent \textbf{Allocating Compute Between Knowledge Acquisition and Rendering.}
We use the fitted laws to derive a compute allocation for the two-stage framework. For total compute $C$, let
$
C=C_{\mathrm{pre}}+C_{\mathrm{post}},
  r=\frac{C_{\mathrm{post}}}{C}.
  % \label{eq:compute_allocation}
$
where $r$ is the fraction of total compute assigned to post-training. Different values of $r$ place different emphasis on knowledge acquisition and rendering.

Autoregressive cross-entropy and diffusion velocity-MSE measure different prediction tasks, so we do not combine them into a single loss. Instead, for each paired AR backbone and renderer, we use the fitted laws to determine each stage's compute-optimal data budget. The corresponding compute costs then specify the allocation $(C_{\mathrm{pre}},C_{\mathrm{post}})$.

\begin{wrapfigure}{r}{0.35\linewidth}
  \centering
  % \vspace{-\baselineskip}
  \vspace{-0.1in}
    \includegraphics[width=0.95\linewidth]{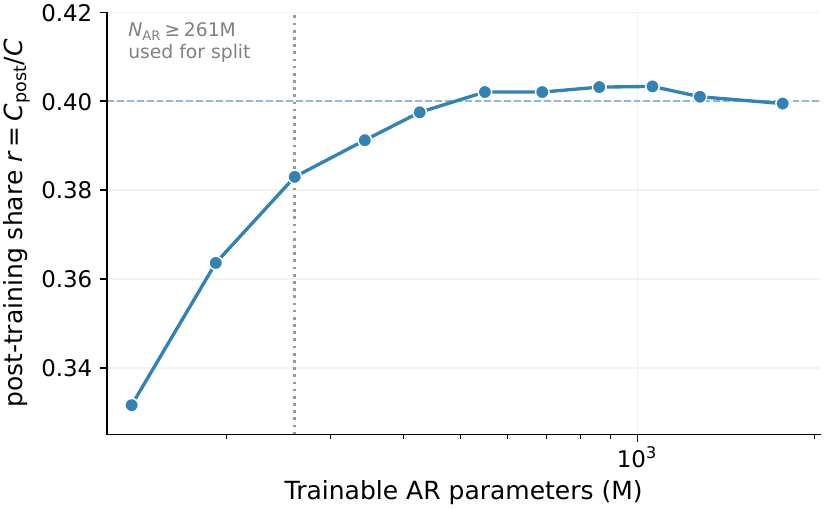}
  \vspace{-0.15in}
  \caption{\small{Compute-optimal post-training share $r$
  against model size.} }
  \label{fig:compute_allocation}
  \vspace{-0.1in}
\end{wrapfigure}

The induced allocation is nearly scale-invariant. Under our primary
full-sequence accounting,$C_{\mathrm{pre}}:C_{\mathrm{post}}\approx3{:}2$
($r\approx0.40$; Fig.~\ref{fig:compute_allocation}). This assigns approximately 60\% of the training budget to knowledge acquisition and 40\% to rendering. Appendix~\ref{app:post_protocol} provides the accounting details.

\noindent \textbf{Scaling recipe for the final model.}
We translate this allocation into a training recipe for the final model. Given a total budget $C$, we assign $C_{\mathrm{pre}}=(1-r)C$ to autoregressive pretraining. The pretraining law then determines the compute-optimal backbone size $N_{\mathrm{AR}}$ and data budget $D_{\mathrm{pre}}$. Within our paired model family, the backbone size also specifies the renderer size. We then use the post-training law to determine its compute-optimal data budget $D_{\mathrm{post}}$. This procedure provides the model configuration and training budgets used for the larger-scale evaluation in the next section.

\vspace{-2mm}
\subsection{Validating the Learning Paradigm at Scale}
\label{sec:exp_hero}
\vspace{-1mm}

Finally, we evaluate whether the training recipe developed through our controlled experiments and scaling analysis remains effective at larger scale. Guided by the scaling laws in Sec.~\ref{sec:exp_scaling}, we train a roughly 3B-parameter text-to-image generator using approximately 6,600 H100 GPU-hours. We evaluate compositional accuracy and text--image alignment on GenEval and DPG-Bench, with results reported in Tab.~\ref{tab:hero_results}. Category-level scores for HiDream-I1-Full, PixNerd, and SeFi-Image-5B are taken from their original reports~\citep{cai2025hidream,wang2026pixnerd,sefi2026image}. For BLIP3o-4B, BLIP3o-NEXT SFT, and i1, we use the aggregate scores reported by \citet{zeng2026i1}, as category-level scores are unavailable under that evaluation.

% Guided by the scaling laws in Sec.~\ref{sec:exp_scaling}, we train an
% approximately 3B-parameter text-to-image generator using approximately 6,600
% H100 GPU-hours. We evaluate the resulting model on GenEval and DPG-Bench,
% which measure compositional accuracy and text--image alignment. Tab.~\ref{tab:hero_results} reports the results.
% The category scores for HiDream-I1-Full, PixNerd, and SeFi-Image-5B follow
% their original reports~\citep{cai2025hidream,wang2026pixnerd,sefi2026image}.
% For BLIP3o-4B, BLIP3o-NEXT SFT, and i1, we use the aggregate results reported by
% \citet{zeng2026i1}; their category scores are not available under that evaluation.

Our model achieves 0.86 on GenEval and 87.78 on DPG-Bench using the evaluation protocol in Appendix~\ref{app:evaluation_protocol}, which includes prompt rewriting and selection over fixed evaluation seeds. Among the baselines in Tab.~\ref{tab:hero_results}, only Qwen-Image and Seedream~3.0 report higher DPG-Bench scores. Our GenEval score is also close to those of Qwen-Image and BLIP3o-NEXT SFT (both 0.87) and SeFi-Image-5B (0.88). These results support the practical effectiveness of our training recipe at larger scale. Comparisons across systems remain approximate because text encoders, training data, and evaluation settings differ. The results demonstrate competitive generation quality within the stated training budget, without establishing that the compute allocation is optimal.

% The model trained with our scaling-informed recipe achieves 0.86 on GenEval
% and 87.78 on DPG-Bench under the evaluation procedure described in
% Appendix~\ref{app:evaluation_protocol}, including prompt rewriting and
% selection over fixed evaluation seeds. Among the reported baselines in
% Tab.~\ref{tab:hero_results}, only Qwen-Image and Seedream~3.0 have higher
% DPG-Bench scores. Its GenEval score is close to those reported for
% Qwen-Image and BLIP3o-NEXT SFT (both 0.87), and SeFi-Image-5B (0.88).
% These results support the practical effectiveness of the recipe at larger
% scale. Differences in text encoders, training data, and evaluation settings
% limit direct attribution and cross-paper comparison; the results do not
% by themselves establish compute optimality.

% \FloatBarrier
\vspace{-1mm}
\section{Related Work}
\label{sec:related_work}
\vspace{-1mm}

\noindent \textbf{Diffusion-based visual generation.}
Diffusion models have evolved from pixel-space denoising~\citep{ho2020ddpm} to
latent diffusion~\citep{rombach2022latentdiffusion}, transformer
backbones~\citep{peebles2023dit}, and flow-based
objectives~\citep{lipman2023flowmatching,liu2023rectifiedflow,
esser2024scalingrectifiedflow}, underpinning modern text-to-image
systems~\citep{bfl2024flux1dev,bfl2025flux2,zimage2025,cai2025hidream}.
Recent work shows that representation quality matters beyond reconstruction
fidelity~\citep{skorokhodov2025diffusability,yao2025vavae}, motivating
semantically enriched tokenizers and diffusion representations~\citep{chen2025maetok,chen2025aligntok,zheng2025rae,tong2026raet2i,
yu2024repa,leng2025repae,singh2025irepa,sefi2026image}.
These approaches make diffusion representations more semantic, but semantic
learning remains coupled to reconstruction or diffusion itself. We instead
separate knowledge acquisition from rendering: autoregressive pretraining learns
the semantic representation consumed by the renderer, while diffusion is
specialized for rendering it into pixels. This separation further enables
Semantic-LM, trained purely through language supervision without an image
decoder or reconstruction loss.

% \vspace{-1.7mm}
\noindent \textbf{Autoregressive visual generation.}
Autoregressive image models predict learned visual tokens~\citep{van2017neural,esser2021taming,yu2023magvitv2,shi2025ibq,
zhao2025bsq,zhao2026spherical} using raster, next-scale, or randomized
prediction~\citep{sun2024llamagen,tian2024var,pang2024randar}.
Tokenizer design is central to this paradigm~\citep{yu2023magvitv2}, with
recent work incorporating foundation-model representations or semantic
supervision~\citep{zheng2025vfmtok,ma2025unitok}. In these approaches,
autoregression remains part of image generation, so its representation must support visual synthesis. We instead specialize autoregression for
knowledge acquisition and delegate rendering to diffusion, freeing its
prediction targets from reconstruction or AR-only generation.

% \vspace{-1.7mm}
\noindent \textbf{Combining autoregression and diffusion.}
Several works combine autoregressive and diffusion modeling within a single
generative framework. MAR~\citep{li2024mar} uses diffusion as the conditional
distribution within continuous-token autoregression, while
Transfusion~\citep{zhou2024transfusion} applies autoregressive and diffusion
objectives to different modalities in a shared model. More closely related,
concurrent VoT~\citep{sun2026vot} predicts semantic visual tokens before
diffusion rendering to extend pretrained VLMs toward unified model. Our distinction is not in how the
two objectives are combined, but in what they are asked to learn. We assign
autoregression to visual knowledge acquisition during pretraining and introduce
diffusion only afterward for rendering, enabling the two roles to be optimized
and scaled separately.

% \vspace{-1.7mm}
\noindent \textbf{Scaling laws for generative models.}
Scaling laws characterize model--data--compute tradeoffs and compute-optimal
training~\citep{kaplan2020scaling,henighan2020scaling,
hoffmann2022computeoptimal}, while transferable parameterizations enable
consistent optimization across scale~\citep{yang2022mup,dey2025completep,
zheng2025ditmup}. Recent work extends scaling analysis to visual
diffusion~\citep{ge2026chimera}. Whereas prior studies primarily scale a single
learning objective, we study cross-objective allocation between knowledge
acquisition and rendering.
% \vspace{-2mm}
\section{Conclusion}
\vspace{-1mm}

We propose a visual generation paradigm that separates knowledge acquisition from rendering: autoregressive pretraining learns visual knowledge, and diffusion renders it into pixels. Through controlled experiments and scaling analysis, we identify three advantages of this approach. First, semantic tokens provide effective pretraining targets without requiring image reconstruction or standalone AR generation. Second, stage-specific scaling laws guide compute allocation, while the lower fitted asymptotic rendering loss suggests benefits of the pretrained pipeline beyond faster convergence. Third, these findings translate into a compute-efficient text-to-image generator that achieves 0.86 on GenEval and 87.78 on DPG-Bench with approximately 6,600 H100 GPU-hours of training. Together, our results support a practical approach to visual generation that tailors the learning objective, representation, and compute budget to each stage.

\bibliographystyle{iclr2027_conference}
\bibliography{references_experiments,iclr2027_conference}

@inproceedings{rombach2022latentdiffusion,
  title = {High-Resolution Image Synthesis with Latent Diffusion Models},
  author = {Rombach, Robin and Blattmann, Andreas and Lorenz, Dominik and
            Esser, Patrick and Ommer, Bjorn},
  booktitle = {Proceedings of the IEEE/CVF Conference on Computer Vision and Pattern Recognition},
  year = {2022},
  url = {https://arxiv.org/abs/2112.10752}
}

@article{yu2023magvitv2,
  title = {Language Model Beats Diffusion---Tokenizer is Key to Visual Generation},
  author = {Yu, Lijun and Lezama, Jose and Gundavarapu, Nitesh B. and
            Versari, Luca and Sohn, Kihyuk and Minnen, David and Cheng, Yong and
            Birodkar, Vighnesh and Gupta, Agrim and Gu, Xiuye and Hauptmann, Alexander G. and
            Gong, Boqing and Yang, Ming-Hsuan and Essa, Irfan and Ross, David A. and Jiang, Lu},
  journal = {arXiv preprint arXiv:2310.05737},
  year = {2023},
  url = {https://arxiv.org/abs/2310.05737}
}

@inproceedings{zhao2025bsq,
  title = {Image and Video Tokenization with Binary Spherical Quantization},
  author = {Zhao, Yue and Xiong, Yuanjun and Kr{\"a}henb{\"u}hl, Philipp},
  booktitle = {International Conference on Learning Representations},
  year = {2025},
  url = {https://arxiv.org/abs/2406.07548}
}

@inproceedings{shi2025ibq,
  title = {Scalable Image Tokenization with Index Backpropagation Quantization},
  author = {Shi, Fengyuan and Luo, Zhuoyan and Ge, Yixiao and Yang, Yujiu and
            Shan, Ying and Wang, Limin},
  booktitle = {Proceedings of the IEEE/CVF International Conference on Computer Vision},
  year = {2025},
  url = {https://arxiv.org/abs/2412.02692}
}

@article{zheng2025rae,
  title = {Diffusion Transformers with Representation Autoencoders},
  author = {Zheng, Boyang and Ma, Nanye and Tong, Shengbang and Xie, Saining},
  journal = {arXiv preprint arXiv:2510.11690},
  year = {2025},
  url = {https://arxiv.org/abs/2510.11690}
}

@inproceedings{ghosh2023geneval,
  title = {{GenEval}: An Object-Focused Framework for Evaluating Text-to-Image Alignment},
  author = {Ghosh, Dhruba and Hajishirzi, Hannaneh and Schmidt, Ludwig},
  booktitle = {Advances in Neural Information Processing Systems},
  year = {2023},
  url = {https://arxiv.org/abs/2310.11513}
}

@article{hu2024ella,
  title = {{ELLA}: Equip Diffusion Models with {LLM} for Enhanced Semantic Alignment},
  author = {Hu, Xiwei and Wang, Rui and Fang, Yixiao and Fu, Bin and Cheng, Pei and Yu, Gang},
  journal = {arXiv preprint arXiv:2403.05135},
  year = {2024},
  url = {https://arxiv.org/abs/2403.05135}
}

@inproceedings{dey2025completep,
  title = {Don't be lazy: {CompleteP} enables compute-efficient deep transformers},
  author = {Dey, Nolan and Zhang, Bin Claire and Noci, Lorenzo and Li, Mufan and
            Bordelon, Blake and Bergsma, Shane and Pehlevan, Cengiz and Hanin, Boris
            and Hestness, Joel},
  booktitle = {Advances in Neural Information Processing Systems},
  year = {2025},
  url = {https://arxiv.org/abs/2505.01618}
}

@article{ge2026chimera,
  title = {{Chimera}: Designing and {Chinchilla}-Scaling Hybrid Visual Diffusion Transformers},
  author = {Ge, Chongjian and Jiang, Hanwen and Wang, Tianyu and Gu, Jiuxiang and
            Xu, Yiran and Chen, Ziwen and Liu, Shaoteng and Shi, Jing and
            Hong, Yicong and Cai, Zefan and Jin, Hailin and Tan, Hao},
  journal = {arXiv preprint arXiv:2607.28611},
  year = {2026},
  url = {https://arxiv.org/abs/2607.28611}
}

@article{hoffmann2022computeoptimal,
  title = {Training Compute-Optimal Large Language Models},
  author = {Hoffmann, Jordan and Borgeaud, Sebastian and Mensch, Arthur and
            Buchatskaya, Elena and Cai, Trevor and Rutherford, Eliza and
            de Las Casas, Diego and Hendricks, Lisa Anne and Welbl, Johannes and
            Clark, Aidan and Hennigan, Tom and Noland, Eric and Millican, Katie and
            van den Driessche, George and Damoc, Bogdan and Guy, Aurelia and
            Osindero, Simon and Simonyan, Karen and Elsen, Erich and
            Rae, Jack W. and Vinyals, Oriol and Sifre, Laurent},
  journal = {arXiv preprint arXiv:2203.15556},
  year = {2022},
  url = {https://arxiv.org/abs/2203.15556}
}

@article{van2017neural,
  title = {Neural discrete representation learning},
  author = {Van Den Oord, Aaron and Vinyals, Oriol and others},
  journal = {Advances in neural information processing systems},
  volume = {30},
  year = {2017}
}

@inproceedings{esser2021taming,
  title = {Taming transformers for high-resolution image synthesis},
  author = {Esser, Patrick and Rombach, Robin and Ommer, Bjorn},
  booktitle = {Proceedings of the IEEE/CVF conference on computer vision and pattern recognition},
  pages = {12873--12883},
  year = {2021}
}

@article{cui2025emu3,
  title = {Emu3.5: Native multimodal models are world learners},
  author = {Cui, Yufeng and Chen, Honghao and Deng, Haoge and Huang, Xu and
            Li, Xinghang and Liu, Jirong and Liu, Yang and Luo, Zhuoyan and
            Wang, Jinsheng and Wang, Wenxuan and others},
  journal = {arXiv preprint arXiv:2510.26583},
  year = {2025}
}

@misc{bai2025qwen25vltechnicalreport,
  title = {Qwen2.5-VL Technical Report},
  author = {Shuai Bai and Keqin Chen and Xuejing Liu and Jialin Wang and
            Wenbin Ge and Sibo Song and Kai Dang and Peng Wang and
            Shijie Wang and Jun Tang and Humen Zhong and Yuanzhi Zhu and
            Mingkun Yang and Zhaohai Li and Jianqiang Wan and Pengfei Wang and
            Wei Ding and Zheren Fu and Yiheng Xu and Jiabo Ye and Xi Zhang and
            Tianbao Xie and Zesen Cheng and Hang Zhang and Zhibo Yang and
            Haiyang Xu and Junyang Lin},
  year = {2025},
  eprint = {2502.13923},
  archivePrefix = {arXiv},
  primaryClass = {cs.CV},
  url = {https://arxiv.org/abs/2502.13923}
}

@article{pang2024randar,
  title = {RandAR: Decoder-only Autoregressive Visual Generation in Random Orders},
  author = {Pang, Ziqi and Zhang, Tianyuan and Luan, Fujun and Man, Yunze and
            Tan, Hao and Zhang, Kai and Freeman, William T. and Wang, Yu-Xiong},
  journal = {arXiv preprint arXiv:2412.01827},
  year = {2024}
}

@article{sun2024llamagen,
  title = {Autoregressive Model Beats Diffusion: Llama for Scalable Image Generation},
  author = {Sun, Peize and Jiang, Yi and Chen, Shoufa and Zhang, Shilong and
            Peng, Bingyue and Luo, Ping and Yuan, Zehuan},
  journal = {arXiv preprint arXiv:2406.06525},
  year = {2024}
}

@inproceedings{lipman2023flowmatching,
  title = {Flow Matching for Generative Modeling},
  author = {Lipman, Yaron and Chen, Ricky T. Q. and Ben-Hamu, Heli and
            Nickel, Maximilian and Le, Matt},
  booktitle = {International Conference on Learning Representations},
  year = {2023},
  url = {https://arxiv.org/abs/2210.02747}
}

@inproceedings{liu2023rectifiedflow,
  title = {Flow Straight and Fast: Learning to Generate and Transfer Data with Rectified Flow},
  author = {Liu, Xingchao and Gong, Chengyue and Liu, Qiang},
  booktitle = {International Conference on Learning Representations},
  year = {2023},
  url = {https://arxiv.org/abs/2209.03003}
}

@inproceedings{peebles2023dit,
  title = {Scalable Diffusion Models with Transformers},
  author = {Peebles, William and Xie, Saining},
  booktitle = {Proceedings of the IEEE/CVF International Conference on Computer Vision},
  year = {2023},
  url = {https://arxiv.org/abs/2212.09748}
}

@article{yang2022mup,
  title = {Tensor Programs {V}: Tuning Large Neural Networks via Zero-Shot Hyperparameter Transfer},
  author = {Yang, Greg and Hu, Edward J. and Babuschkin, Igor and Sidor, Szymon and
            Liu, Xiaodong and Farhi, David and Ryder, Nick and Pachocki, Jakub and
            Chen, Weizhu and Gao, Jianfeng},
  journal = {arXiv preprint arXiv:2203.03466},
  year = {2022}
}

@article{farina2026datacomp,
  title = {{DataComp-VLM}: Improved Open Datasets for Vision-Language Models},
  author = {Farina, Matteo and Udandarao, Vishaal and Nguyen, Thao and
            Kuzucu, Selim and B{\"o}ther, Maximilian and Hochlehnert, Andreas and
            Ghosh, Adhiraj and Nezhurina, Marianna and Roth, Karsten and
            Struber, Joschka and Zhang, Yuhui and Dziadzio, Sebastian and
            Sui, Elaine and Jahagirdar, Soumya and Ghosh, Dhruba and
            Hammoud, Hasan and De Min, Thomas and Caldarella, Simone and
            Mirza, Jehanzeb and Keh, Sedrick and Cherti, Mehdi and
            Kuehne, Hilde and Schiele, Bernt and Yeung-Levy, Serena and
            Naeem, Muhammad Ferjad and Tombari, Federico and Klimovic, Ana and
            Ricci, Elisa and Bethge, Matthias and Oh, Sewoong and
            Prabhu, Ameya and Tonioni, Alessio and Jitsev, Jenia and
            Mancini, Massimiliano and Schmidt, Ludwig and Parthasarathy, Nikhil},
  journal = {arXiv preprint arXiv:2606.28551},
  year = {2026},
  url = {https://arxiv.org/abs/2606.28551}
}

@article{shi2025svgt2i,
  title = {{SVG-T2I}: Scaling Up Text-to-Image Latent Diffusion Model Without Variational Autoencoder},
  author = {Shi, Minglei and Wang, Haolin and Zhang, Borui and Zheng, Wenzhao and
            Zeng, Bohan and Yuan, Ziyang and Wu, Xiaoshi and Zhang, Yuanxing and
            Yang, Huan and Wang, Xintao and Wan, Pengfei and Gai, Kun and
            Zhou, Jie and Lu, Jiwen},
  journal = {arXiv preprint arXiv:2512.11749},
  year = {2025},
  url = {https://arxiv.org/abs/2512.11749}
}

@article{simeoni2025dinov3,
  title = {{DINOv3}},
  author = {Sim{\'e}oni, Oriane and Vo, Huy V. and Seitzer, Maximilian and
            Baldassarre, Federico and Oquab, Maxime and others},
  journal = {arXiv preprint arXiv:2508.10104},
  year = {2025},
  url = {https://arxiv.org/abs/2508.10104}
}

@misc{bfl2025flux2,
  title = {{FLUX.2}: Official Inference Repository and Autoencoder},
  author = {{Black Forest Labs}},
  year = {2025},
  howpublished = {\url{https://github.com/black-forest-labs/flux2}}
}

@article{tian2024var,
  title = {Visual Autoregressive Modeling: Scalable Image Generation via Next-Scale Prediction},
  author = {Tian, Keyu and Jiang, Yi and Yuan, Zehuan and Peng, Bingyue and Wang, Liwei},
  journal = {arXiv preprint arXiv:2404.02905},
  year = {2024},
  url = {https://arxiv.org/abs/2404.02905}
}

@article{kaplan2020scaling,
  title = {Scaling Laws for Neural Language Models},
  author = {Kaplan, Jared and McCandlish, Sam and Henighan, Tom and Brown, Tom B.
            and Chess, Benjamin and Child, Rewon and Gray, Scott and Radford, Alec
            and Wu, Jeffrey and Amodei, Dario},
  journal = {arXiv preprint arXiv:2001.08361},
  year = {2020},
  url = {https://arxiv.org/abs/2001.08361}
}

@article{esser2024scalingrectifiedflow,
  title = {Scaling Rectified Flow Transformers for High-Resolution Image Synthesis},
  author = {Esser, Patrick and Kulal, Sumith and Blattmann, Andreas and Entezari, Rahim and M{\"u}ller, Jonas and Saini, Harry and Levi, Yam and Lorenz, Dominik and Sauer, Axel and Boesel, Frederic and Podell, Dustin and Dockhorn, Tim and English, Zion and Lacey, Kyle and Goodwin, Alex and Marek, Yannik and Rombach, Robin},
  journal = {arXiv preprint arXiv:2403.03206},
  year = {2024},
  url = {https://arxiv.org/abs/2403.03206}
}

@article{ho2020ddpm,
  title = {Denoising Diffusion Probabilistic Models},
  author = {Ho, Jonathan and Jain, Ajay and Abbeel, Pieter},
  journal = {arXiv preprint arXiv:2006.11239},
  year = {2020},
  url = {https://arxiv.org/abs/2006.11239}
}

@misc{bfl2024flux1dev,
  title = {{FLUX.1 [dev]}},
  author = {{Black Forest Labs}},
  year = {2024},
  howpublished = {\url{https://huggingface.co/black-forest-labs/FLUX.1-dev}}
}

@article{zimage2025,
  title = {{Z-Image}: An Efficient Image Generation Foundation Model with Single-Stream Diffusion Transformer},
  author = {{Z-Image Team} and Cai, Huanqia and Cao, Sihan and Du, Ruoyi and others},
  journal = {arXiv preprint arXiv:2511.22699},
  year = {2025},
  url = {https://arxiv.org/abs/2511.22699}
}

@article{cai2025hidream,
  title = {{HiDream-I1}: A High-Efficient Image Generative Foundation Model with Sparse Diffusion Transformer},
  author = {Cai, Qi and Chen, Jingwen and Chen, Yang and others},
  journal = {arXiv preprint arXiv:2505.22705},
  year = {2025},
  url = {https://arxiv.org/abs/2505.22705}
}

@article{wang2026pixnerd,
  title = {{PixNerd}: Pixel Neural Field Diffusion},
  author = {Wang, Shuai and Gao, Ziteng and Zhu, Chenhui and Huang, Weilin and Wang, Limin},
  journal = {International Conference on Learning Representations},
  year = {2026},
  url = {https://arxiv.org/abs/2507.23268}
}

@article{zeng2026i1,
  title = {{i1}: A Simple and Fully Open Recipe for Strong Text-to-Image Models},
  author = {Zeng, Boya and Luo, Tianze and Pu, Shu and Shen, Jucheng and Lu, Taiming and Sarch, Gabriel and Liu, Zhuang},
  journal = {arXiv preprint arXiv:2606.11289},
  year = {2026},
  url = {https://arxiv.org/abs/2606.11289}
}

@article{sefi2026image,
  title = {{SeFi-Image}: A Text-to-Image Foundation Model with Semantic-First Diffusion},
  author = {{SeFi Team}},
  journal = {arXiv preprint arXiv:2606.22568},
  year = {2026},
  url = {https://arxiv.org/abs/2606.22568}
}

@misc{openai2026gptimage2,
  author = {{OpenAI}},
  title  = {{GPT Image 2}},
  year   = {2026},
  url    = {https://developers.openai.com/api/docs/models/gpt-image-2}
}

@misc{ideogram4,
  author = {{Ideogram}},
  title  = {{Ideogram 4.0}},
  url    = {https://ideogram.ai/models/4.0/},
  note   = {Accessed September 23, 2026}
}

@misc{reve2026,
  author = {{Reve}},
  title  = {{Reve 2.1}},
  year   = {2026},
  url    = {https://www.reve.com/model}
}

@article{chen2026scaling,
  author  = {Chen, Zilong and Deng, Chaorui and Li, Kunchang
             and Yuan, Hongyi and Fan, Haoqi},
  title   = {Scaling Properties of Text Conditioning in Visual Generation},
  journal = {arXiv preprint arXiv:2607.29679},
  year    = {2026},
  doi     = {10.48550/arXiv.2607.29679},
  url     = {https://arxiv.org/abs/2607.29679}
}

@inproceedings{zhao2026spherical,
  title={Spherical leech quantization for visual tokenization and generation},
  author={Zhao, Yue and Jiang, Hanwen and Xu, Zhenlin and Yang, Chutong and Adeli, Ehsan and Kr{\"a}henb{\"u}hl, Philipp},
  booktitle={Proceedings of the IEEE/CVF Conference on Computer Vision and Pattern Recognition},
  pages={12913--12923},
  year={2026}
}

@article{yu2024repa,
  title = {{Representation Alignment for Generation: Training Diffusion Transformers Is Easier Than You Think}},
  author = {Yu, Sihyun and Kwak, Sangkyung and Jang, Huiwon and Jeong, Jongheon and Huang, Jonathan and Shin, Jinwoo and Xie, Saining},
  journal = {arXiv preprint arXiv:2410.06940},
  year = {2024},
  url = {https://arxiv.org/abs/2410.06940}
}

@article{li2024mar,
  title = {{Autoregressive Image Generation without Vector Quantization}},
  author = {Li, Tianhong and Tian, Yonglong and Li, He and Deng, Mingyang and He, Kaiming},
  journal = {arXiv preprint arXiv:2406.11838},
  year = {2024},
  url = {https://arxiv.org/abs/2406.11838}
}

@article{zhou2024transfusion,
  title = {{Transfusion: Predict the Next Token and Diffuse Images with One Multi-Modal Model}},
  author = {Zhou, Chunting and Yu, Lili and Babu, Arun and Tirumala, Kushal and Yasunaga, Michihiro and Shamis, Leonid and Kahn, Jacob and Ma, Xuezhe and Zettlemoyer, Luke and Levy, Omer},
  journal = {arXiv preprint arXiv:2408.11039},
  year = {2024},
  url = {https://arxiv.org/abs/2408.11039}
}

@article{henighan2020scaling,
  title = {{Scaling Laws for Autoregressive Generative Modeling}},
  author = {Henighan, Tom and Kaplan, Jared and Katz, Mor and Chen, Mark and Hesse, Christopher and Jackson, Jacob and Jun, Heewoo and Brown, Tom B. and Dhariwal, Prafulla and Gray, Scott and Hallacy, Chris and Mann, Benjamin and Radford, Alec and Ramesh, Aditya and Ryder, Nick and Ziegler, Daniel M. and Schulman, John and Amodei, Dario and McCandlish, Sam},
  journal = {arXiv preprint arXiv:2010.14701},
  year = {2020},
  url = {https://arxiv.org/abs/2010.14701}
}

@article{chen2025aligntok,
  title = {{AlignTok: Aligning Visual Foundation Encoders to Tokenizers for Diffusion Models}},
  author = {Chen, Bowei and Bi, Sai and Tan, Hao and Zhang, He and Zhang, Tianyuan and Li, Zhengqi and Xiong, Yuanjun and Zhang, Jianming and Zhang, Kai},
  journal = {arXiv preprint arXiv:2509.25162},
  year = {2025},
  url = {https://arxiv.org/abs/2509.25162}
}

@article{yao2025vavae,
  title = {{Reconstruction vs. Generation: Taming Optimization Dilemma in Latent Diffusion Models}},
  author = {Yao, Jingfeng and Yang, Bin and Wang, Xinggang},
  journal = {arXiv preprint arXiv:2501.01423},
  year = {2025},
  url = {https://arxiv.org/abs/2501.01423}
}

@article{chen2025maetok,
  title = {{Masked Autoencoders Are Effective Tokenizers for Diffusion Models}},
  author = {Chen, Hao and Han, Yujin and Chen, Fangyi and Li, Xiang and Wang, Yidong and Wang, Jindong and Wang, Ze and Liu, Zicheng and Zou, Difan and Raj, Bhiksha},
  journal = {arXiv preprint arXiv:2502.03444},
  year = {2025},
  url = {https://arxiv.org/abs/2502.03444}
}

@article{leng2025repae,
  title = {{REPA-E: Unlocking VAE for End-to-End Tuning with Latent Diffusion Transformers}},
  author = {Leng, Xingjian and Singh, Jaskirat and Hou, Yunzhong and Xing, Zhenchang and Xie, Saining and Zheng, Liang},
  journal = {arXiv preprint arXiv:2504.10483},
  year = {2025},
  url = {https://arxiv.org/abs/2504.10483}
}

@article{zheng2025vfmtok,
  title = {{Vision Foundation Models as Effective Visual Tokenizers for Autoregressive Image Generation}},
  author = {Zheng, Anlin and Wen, Xin and Zhang, Xuanyang and Ma, Chuofan and Wang, Tiancai and Yu, Gang and Zhang, Xiangyu and Qi, Xiaojuan},
  journal = {arXiv preprint arXiv:2507.08441},
  year = {2025},
  url = {https://arxiv.org/abs/2507.08441}
}

@article{ma2025unitok,
  title = {{UniTok: A Unified Tokenizer for Visual Generation and Understanding}},
  author = {Ma, Chuofan and Jiang, Yi and Wu, Junfeng and Yang, Jihan and Yu, Xin and Yuan, Zehuan and Peng, Bingyue and Qi, Xiaojuan},
  journal = {arXiv preprint arXiv:2502.20321},
  year = {2025},
  url = {https://arxiv.org/abs/2502.20321}
}

@article{tong2026raet2i,
  title = {{Scaling Text-to-Image Diffusion Transformers with Representation Autoencoders}},
  author = {Tong, Shengbang and Zheng, Boyang and Wang, Ziteng and Tang, Bingda and Ma, Nanye and Brown, Ellis and Yang, Jihan and Fergus, Rob and LeCun, Yann and Xie, Saining},
  journal = {arXiv preprint arXiv:2601.16208},
  year = {2026},
  url = {https://arxiv.org/abs/2601.16208}
}

@article{skorokhodov2025diffusability,
  title = {{Improving the Diffusability of Autoencoders}},
  author = {Skorokhodov, Ivan and Girish, Sharath and Hu, Benran and Menapace, Willi and Li, Yanyu and Abdal, Rameen and Tulyakov, Sergey and Siarohin, Aliaksandr},
  journal = {arXiv preprint arXiv:2502.14831},
  year = {2025},
  url = {https://arxiv.org/abs/2502.14831}
}

@article{singh2025irepa,
  title = {{What matters for Representation Alignment: Global Information or Spatial Structure?}},
  author = {Singh, Jaskirat and Leng, Xingjian and Wu, Zongze and Zheng, Liang and Zhang, Richard and Shechtman, Eli and Xie, Saining},
  journal = {arXiv preprint arXiv:2512.10794},
  year = {2025},
  url = {https://arxiv.org/abs/2512.10794}
}

@article{zheng2025ditmup,
  title = {{Scaling Diffusion Transformers Efficiently via $\mu$P}},
  author = {Zheng, Chenyu and Zhang, Xinyu and Wang, Rongzhen and Huang, Wei and Tian, Zhi and Huang, Weilin and Zhu, Jun and Li, Chongxuan},
  journal = {arXiv preprint arXiv:2505.15270},
  year = {2025},
  url = {https://arxiv.org/abs/2505.15270}
}

@article{sun2026vot,
  title = {{VoT: Vision-of-Thought for Unified Multimodal Representation Alignment}},
  author = {Sun, Jingxiang and Liao, Chao and Luo, Zhengxiong and Deng, Chaorui and Zhang, Chen-lin and Wang, Junke and Yang, Ceyuan and Fan, Haoqi and Huang, Weilin},
  journal = {arXiv preprint arXiv:2609.07815},
  year = {2026},
  url = {https://arxiv.org/abs/2609.07815}
}

@article{chen2023pixart,
  title = {{PixArt-$\alpha$}: Fast Training of Diffusion Transformer for Photorealistic Text-to-Image Synthesis},
  author = {Chen, Junsong and Yu, Jincheng and Ge, Chongjian and Yao, Lewei and Xie, Enze and Wu, Yue and Wang, Zhongdao and Kwok, James and Luo, Ping and Lu, Huchuan and Li, Zhenguo},
  journal = {arXiv preprint arXiv:2310.00426},
  year = {2023},
  url = {https://arxiv.org/abs/2310.00426}
}

@article{podell2023sdxl,
  title = {{SDXL}: Improving Latent Diffusion Models for High-Resolution Image Synthesis},
  author = {Podell, Dustin and English, Zion and Lacey, Kyle and Blattmann, Andreas and Dockhorn, Tim and M{\"u}ller, Jonas and Penna, Joe and Rombach, Robin},
  journal = {arXiv preprint arXiv:2307.01952},
  year = {2023},
  url = {https://arxiv.org/abs/2307.01952}
}

@article{chen2025blip3o,
  title = {{BLIP3-o}: A Family of Fully Open Unified Multimodal Models---Architecture, Training and Dataset},
  author = {Chen, Jiuhai and Xu, Zhiyang and Pan, Xichen and Hu, Yushi and Qin, Can and Goldstein, Tom and Huang, Lifu and Zhou, Tianyi and Xie, Saining and Savarese, Silvio and Xue, Le and Xiong, Caiming and Xu, Ran},
  journal = {arXiv preprint arXiv:2505.09568},
  year = {2025},
  url = {https://arxiv.org/abs/2505.09568}
}

@article{gao2025seedream3,
  title = {{Seedream 3.0} Technical Report},
  author = {Gao, Yu and Gong, Lixue and Guo, Qiushan and others},
  journal = {arXiv preprint arXiv:2504.11346},
  year = {2025},
  url = {https://arxiv.org/abs/2504.11346}
}

@article{wu2025qwenimage,
  title = {{Qwen-Image} Technical Report},
  author = {Wu, Chenfei and Li, Jiahao and Zhou, Jingren and others},
  journal = {arXiv preprint arXiv:2508.02324},
  year = {2025},
  url = {https://arxiv.org/abs/2508.02324}
}

@article{chen2025blip3onext,
  title = {{BLIP3o-NEXT}: Next Frontier of Native Image Generation},
  author = {Chen, Jiuhai and Xue, Le and Xu, Zhiyang and Pan, Xichen and Yang, Shusheng and Qin, Can and Yan, An and Zhou, Honglu and Chen, Zeyuan and Huang, Lifu and Zhou, Tianyi and Li, Junnan and Savarese, Silvio and Xiong, Caiming and Xu, Ran},
  journal = {arXiv preprint arXiv:2510.15857},
  year = {2025},
  url = {https://arxiv.org/abs/2510.15857}
}

@misc{betker2023dalle3,
  title = {Improving Image Generation with Better Captions},
  author = {Betker, James and Goh, Gabriel and Jing, Li and Brooks, Tim and Wang, Jianfeng and Li, Linjie and Ouyang, Long and Zhuang, Juntang and Lee, Joyce and Guo, Yufei and Manassra, Wesam and Dhariwal, Prafulla and Chu, Casey and Jiao, Yunxin and Ramesh, Aditya},
  year = {2023},
  howpublished = {OpenAI technical report},
  url = {https://cdn.openai.com/papers/dall-e-3.pdf}
}

\clearpage
\appendix
\section{Additional Experimental Details}
\label{app:exp_protocol}

We provide training and evaluation details for the controlled comparisons,
hyperparameter studies, and scaling experiments.

\subsection{Semantic Tokenizer Training}
\label{app:tokenizer_training}

We train two semantic tokenizers to study the role of codebook supervision:
Semantic-Recon uses image reconstruction, whereas Semantic-LM uses language
modeling. Both quantize post-merger features from the frozen Qwen2.5-VL-32B-Instruct
visual encoder using an
IBQ codebook~\citep{shi2025ibq}. Tokenizer training is separate from the
AR-pretraining and diffusion-post-training budgets reported in
Appendix~\ref{app:tokenizer_controls}; the resulting tokenizers remain frozen
throughout generator training.

\paragraph{Shared configuration.}
The IBQ codebook contains 131,072 entries with code dimension 256 and
quantization temperature 0.25. Both tokenizers use an entropy regularization
coefficient of 0.05 and a global batch size of 256. The optimizer uses
$\beta_1=0.9$, $\beta_2=0.95$, $\epsilon=10^{-8}$, and weight decay 0.01;
gradients are clipped at norm 1. We linearly warm up the learning rate to
$2\times10^{-4}$ over 10,000 updates, then apply cosine decay with a minimum learning rate of
$2\times10^{-5}$. Training uses three
native-resolution tiers with image areas near $256^2$, $512^2$, and $1024^2$
pixels and random aspect ratios. We sample these tiers in a 4:2:1 ratio
by number of examples. This sample ratio is distinct from the token quotas
used for generator training.

\paragraph{Semantic-Recon.}
An IBQ codebook quantizes the frozen post-merger visual features, and a transformer decoder reconstructs the input
image from the quantized representation. The decoder has 12 layers, hidden
width 768, and 12 attention heads. We train the codebook and decoder on
in-house training data for 80,000 updates. The objective combines image
reconstruction, VGG-based LPIPS perceptual supervision, and quantization
loss, with respective weights 1, 0.5, and 1, together with the entropy
regularizer specified above. The pretrained visual feature extractor remains
frozen. We maintain an exponential moving average of the trainable parameters
with decay 0.9999 and a 2,000-update warmup.

\paragraph{Semantic-LM.}
We freeze Qwen2.5-VL-32B-Instruct, including its visual feature extractor and
language model, and train only the IBQ codebook. Quantized visual features
condition the frozen language model, and next-token cross-entropy on the
supervised text targets provides the codebook's training signal. This
language-modeling objective is not restricted to image captions. We train
for 40,000 updates on the 200B version of
DataComp-VLM~\citep{farina2026datacomp}, using the shared optimization and
resolution settings above. The objective combines language-modeling
cross-entropy with entropy regularization at weight 0.05. No image decoder
or image reconstruction loss is used.

\subsection{Controlled Generation Comparisons}
\label{app:tokenizer_controls}
\label{app:e7_protocol}

\paragraph{Model capacity.}
For the direct-pixel and AR controls, we match the total parameter count
of the final generators at approximately
400M total parameters. The PDM has width 1408, 16 layers,
11 attention heads with head dimension 128, and feed-forward hidden dimension
3840. The AR backbone uses the same width, attention configuration, and
feed-forward dimension, but only eight layers. Diffusion post-training adds
randomly initialized rendering modules, including separate diffusion
feed-forward layers, while keeping the AR parameters frozen. The AR-pretraining
backbone has approximately 250M parameters, and the rendering modules add
approximately 150M, giving a final generator of approximately 400M that matches
the PDM. This matched size refers to the final
generator, not to the number of parameters used during pretraining. Total parameter counts include the
frozen AR backbone and trainable rendering modules, count shared weights
once, and exclude external frozen text encoders and tokenizers.

\paragraph{Training budget and resolution curriculum.}
Each generator in the original direct-pixel and AR control receives
400,000 optimizer updates. PDM
uses its diffusion objective throughout. The pretraining--post-training models use 150,000
AR-pretraining updates followed by 250,000 diffusion-post-training updates.

\begin{table}[!t]
  \centering
  \caption{\textbf{Training recipes for the controlled generation comparisons.}
  All routes match cumulative image exposure and use approximately 400M final
  parameters. Pretraining--post-training budgets include both stages; all AR tokenizer variants
  share the AR$\xrightarrow{}$PDM recipe.}
  \label{tab:experiment_recipe}
  \footnotesize
  \setlength{\tabcolsep}{5pt}
  \renewcommand{\arraystretch}{1.12}
  \begin{tabularx}{\linewidth}{@{}>{\raggedright\arraybackslash}p{0.19\linewidth}
    *{4}{>{\raggedright\arraybackslash}X}@{}}
    \toprule
    \textbf{Setting} & \textbf{PDM} & \textbf{LDM} & \textbf{LDM$\xrightarrow{}$PDM} & \textbf{AR$\xrightarrow{}$PDM} \\
    \midrule
    \multicolumn{5}{@{}l}{\textit{Architecture and representation}}\\
    \addlinespace[3pt]
    Backbone width & 1408 & 1280 & 1280 & 1408 \\
    Layers / heads & 16 / 11 & 15 / 10 & 8 / 10 & 8 / 11 \\
    FFN width & 3840 & 3584 & 3584 & 3840 \\
    \addlinespace[4pt]
    Representation & Pixels & FLUX.2 & SVG Autoencoder-P & Three tokenizers \\
    Latent channels & --- & 128 (packed) & 384 & Discrete codes \\
    Parameters (M) & $\sim$400 & $\sim$400 & $\sim$250 + 150 & $\sim$250 + 150 \\
    Image output & Pixels & Frozen decoder & Pixel renderer & Pixel renderer \\
    \addlinespace[3pt]
    \midrule
    \multicolumn{5}{@{}l}{\textit{Training budget and curriculum}}\\
    \addlinespace[3pt]
    Updates & 400k PDM & 400k LDM & 150k LDM + 250k PDM & 150k AR + 250k PDM \\
    \addlinespace[4pt]
    Resolution tiers & \multicolumn{4}{@{}p{0.76\linewidth}@{}}{Native images with random aspect ratios; areas near $256^2$ and $512^2$.}\\
    Curriculum & \multicolumn{4}{@{}p{0.76\linewidth}@{}}{First 50k updates: low resolution only. Remaining 350k: mixed resolution, counting continuously across stages.}\\
    Token budget & \multicolumn{4}{@{}p{0.76\linewidth}@{}}{262,400 per update; mixed-phase low:high token quotas of 1:4.}\\
    \addlinespace[3pt]
    \midrule
    \multicolumn{5}{@{}l}{\textit{Optimization}}\\
    \addlinespace[3pt]
    Base learning rate & $10^{-3}$ & $10^{-3}$ & SVG-T2I / $6\times10^{-4}$ & $7\times10^{-5}$ / $6\times10^{-4}$ \\
    Warmup updates & 2,000 & 2,000 & SVG-T2I / 0 & 2,000 / 0 \\
    \addlinespace[4pt]
    AdamW & \multicolumn{4}{@{}p{0.76\linewidth}@{}}{$\beta_1=\beta_2=0.95$, $\epsilon=10^{-8}$, weight decay 0.01; gradient clipping at norm 1.}\\
    Schedule / EMA & \multicolumn{4}{@{}p{0.76\linewidth}@{}}{Constant learning rate after warmup; EMA disabled. SVG latent pretraining follows SVG-T2I optimization.}\\
    \addlinespace[3pt]
    \midrule
    \multicolumn{5}{@{}l}{\textit{Diffusion settings}}\\
    \addlinespace[3pt]
    Gaussian noise std. & 2 & 1 & 1 / 2 & 2 (post-training) \\
    Latent recipe & --- & SD3 & SVG-T2I & --- \\
    \addlinespace[4pt]
    Objective & \multicolumn{4}{@{}p{0.76\linewidth}@{}}{Velocity flow matching for all diffusion stages.}\\
    Pixel rendering & \multicolumn{4}{@{}p{0.76\linewidth}@{}}{16-pixel patches; logit-normal timesteps ($\mu=0$, $\sigma=1$); high-resolution shift $s=\sqrt{L_{\mathrm{diff}}/256}$. Both pretraining--post-training routes use the same renderer training recipe.}\\
    \addlinespace[4pt]
    \midrule
    \multicolumn{5}{@{}p{\linewidth}@{}}{\textit{Notes.} Paired optimization and noise entries denote pretraining / post-training; paired parameter counts denote backbone + renderer. The three AR tokenizers are Recon, Sem-Recon, and Sem-LM.}\\
    \addlinespace[3pt]
    \multicolumn{5}{@{}p{\linewidth}@{}}{Latent generators are trained from scratch with frozen released tokenizers: FLUX.2~\citep{bfl2025flux2} and DINOv3-based SVG Autoencoder-P~\citep{shi2025svgt2i,simeoni2025dinov3}. Both retain released preprocessing and normalization; FLUX.2 uses posterior-mean encoding. Latent recipes follow SD3~\citep{esser2024scalingrectifiedflow} or SVG-T2I~\citep{shi2025svgt2i}, as indicated.}\\
    \bottomrule
  \end{tabularx}
\end{table}

We use two native-resolution tiers with random aspect ratios. The low-resolution
tier has image area $HW\approx256^2$, and the high-resolution tier has
image area $HW\approx512^2$; these denote approximate areas rather than
fixed square image shapes. The curriculum follows the cumulative update
count across both stages. The first 50,000 updates use only the low-resolution
tier, and the remaining 350,000 mix the two tiers. Thus AR pretraining
comprises 50,000 low-resolution and 100,000 mixed-resolution updates;
all 250,000 post-training updates use the mixed-resolution phase.

The training token budget is 262,400 per optimizer update throughout.
Within the mixed-resolution phase, the low- and high-resolution
tiers receive token quotas in a 1:4 ratio, corresponding to 52,480 and
209,920 tokens per update. This ratio specifies token allocation, not image
counts or the number of updates assigned to each tier. We align cumulative
image presentations across the direct-pixel and three AR generators
(Tabs.~\ref{tab:efficiency_results} and~\ref{tab:tokenizer_results}),
including both stages for the AR-pretrained models:
\begin{equation}
  D_{\mathrm{img}}^{\mathrm{diffusion}}
  = D_{\mathrm{img}}^{\mathrm{AR}} + D_{\mathrm{img}}^{\mathrm{post}}.
  \label{eq:e7_data_budget}
\end{equation}
We use a deterministic dataloader with the same sample order across
controlled comparisons. With matched training steps and global batch size,
each model sees the same training samples, counting both stages for the
pretraining--post-training models. Repeated presentations count separately.

\paragraph{Tokenizer controls.}
For the comparisons among Recon, Semantic-Recon, and Semantic-LM, the
tokenizer is the only experimental variable. All three use the same generator
architecture, training images and captions, preprocessing, sample order,
resolution curriculum, token budget, and stage durations. We also align
code-embedding initialization, optimization, text conditioning, checkpoint
selection, and generation evaluation. Each tokenizer remains frozen during
generator training. Differences in the tokenizer's visual representation and
codebook supervision are the interventions defined in
Sec.~\ref{sec:method}. Appendix~\ref{app:tokenizer_training} describes
training of the semantic tokenizers, which precedes generator training.
We evaluate reconstruction FID (rFID) on the 50,000 ImageNet validation images.
Semantic-LM has no image decoder, so reconstruction quality is undefined;
Tab.~\ref{tab:tokenizer_results} denotes this case by $\infty$.
For AR-only generation, we evaluate the checkpoint after 150,000
pretraining updates using the same GenEval prompts as the final model and
AR classifier-free guidance scale 3.

\paragraph{Shared diffusion objective and noise schedule.}
PDM and diffusion post-training use the same noising,
resolution shift, and loss convention. Images are normalized to $[-1,1]$.
Using the implementation's clean-time coordinate $u$, we sample
$z\sim\mathcal{N}(\mu,\sigma^2)$ with $\mu=0$ and $\sigma=1$, and set
$u=\operatorname{sigmoid}(z)$. Gaussian noise has standard deviation 2.
For an image with $L_{\mathrm{diff}}$ diffusion patches, the high-resolution
schedule uses
\begin{equation}
  s=\sqrt{L_{\mathrm{diff}}/256},\qquad
  \widetilde u=\frac{u}{u+s(1-u)}.
  \label{eq:comparison_resolution_shift}
\end{equation}
Low-resolution images use $s=1$, including throughout the first 50,000
updates; high-resolution images in the mixed phase use
Eq.~\ref{eq:comparison_resolution_shift}. The diffusion patch size
is 16 pixels, following the reference diffusion recipe. The patch count in
Eq.~\ref{eq:comparison_resolution_shift} is distinct from the AR
code sequence length and the global training token budget.

The implementation forms $x_{\widetilde u}=\widetilde u x_0+
(1-\widetilde u)\epsilon$, where $\epsilon\sim\mathcal{N}(0,4I)$.
The image prediction $\widehat x_0$ is converted to a velocity prediction
using the stabilized denominator $d=\max(1-\widetilde u,0.05)$:
\begin{equation}
  \widehat v=\frac{\widehat x_0-x_{\widetilde u}}{d},\qquad
  v_{\mathrm{target}}=\frac{x_0-x_{\widetilde u}}{d}.
  \label{eq:comparison_velocity}
\end{equation}
We minimize squared velocity error, averaging over RGB values within each
image and then over images. Unequal per-rank image counts are weighted to
recover this global image average. The clean-time coordinate is the reverse
of the noise-time coordinate in Sec.~\ref{sec:method}:
$t=1-\widetilde u$; reversing both velocity signs leaves squared error unchanged.

\paragraph{Optimization.}
We use stage-specific settings selected through prior hyperparameter searches
and share the selected settings across tokenizer variants. The PDM base
learning rate is $10^{-3}$, with 2,000 linear-warmup updates followed by a
constant rate. AR pretraining uses base learning rate $7\times10^{-5}$
and the same warmup duration. Diffusion post-training uses a separately
selected base learning rate of $6\times10^{-4}$, held constant without warmup.
All stages use AdamW with $\beta_1=\beta_2=0.95$, $\epsilon=10^{-8}$,
weight decay 0.01, and gradient clipping at norm 1. EMA is disabled.

\subsection{Pretraining Hyperparameter Transfer}
\label{app:ar_transfer}

\paragraph{Models and frozen representations.}
We vary width at 16 layers and depth at width 512, yielding the 11 distinct
architectures in Tab.~\ref{tab:ar_transfer_models}. Each model is a causal
RandAR transformer with pre-RMSNorm, QK normalization, SwiGLU
feed-forward layers, and axial two-dimensional rotary position embeddings.
The attention head dimension is fixed at 128. The FFN hidden dimension is
$F=256\lceil\lfloor 8W/3\rfloor/256\rceil$, where $W$ is model width.
The input and output share a trainable, factorized code table with
131,072 entries of dimension 256.

All sweep arms share frozen FLAN-T5-XL text features (dimension 2048, maximum
length 512) and the frozen Semantic-Recon tokenizer described in
Appendix~\ref{app:tokenizer_training}. Only the AR generator is optimized.

\begin{table}[htbp]
  \centering
  \caption{Architectures used for AR hyperparameter transfer. Each row is
  trained at all 12 base learning rates. The reference architecture appears
  in both panels of Fig.~\ref{fig:ar_hp_transfer}. Parameter counts include
  all trainable AR parameters, count the tied code table once, and exclude
  the frozen visual tokenizer and text encoder. FFN denotes feed-forward
  hidden dimension; each attention head has dimension 128. Parameter counts
  are rounded to one decimal place in millions (M).}
  \label{tab:ar_transfer_models}
  \small
  \renewcommand{\arraystretch}{1.15}
  \begin{tabular}{@{}lrrrrr@{}}
    \toprule
    Sweep & Width & Layers & Heads & FFN & Parameters \\
    \midrule
    Depth & 512 & 4  & 4 & 1536 & 48.9M \\
          & 512 & 8  & 4 & 1536 & 62.5M \\
          & 512 & 12 & 4 & 1536 & 76.2M \\
          & 512 & 24 & 4 & 1536 & 117.1M \\
          & 512 & 32 & 4 & 1536 & 144.4M \\
    \midrule
    Shared & 512 & 16 & 4 & 1536 & 89.8M \\
    \midrule
    Width & 256  & 16 & 2 & 768  & 48.1M \\
          & 384  & 16 & 3 & 1024 & 63.1M \\
          & 640  & 16 & 5 & 1792 & 117.0M \\
          & 768  & 16 & 6 & 2048 & 149.5M \\
          & 1024 & 16 & 8 & 2816 & 242.9M \\
    \bottomrule
  \end{tabular}
\end{table}

\paragraph{Initialization and transfer rules.}
Our reference dimensions are width $W_0=512$, depth $L_0=4$, and FFN hidden
dimension $F_0=1536$. Define $m_W=W/W_0$ and $m_L=L/L_0$. We fix the base
initialization scale at $s_0=0.02$ throughout the sweep. Attention and FFN
matrices use zero-mean truncated-normal initialization with scale
$s_0/\sqrt{m_W}$. The width-to-width text projection uses a normal
distribution with the same scale. Input/output code projections, the first
text projection, and position-instruction token embeddings use normal
initialization with scale $s_0$; normalization gains start at one and the
output bias at zero. The tied code table is initialized from the frozen
tokenizer's codebook values, then trained as part of the AR model.

Following CompleteP-style depth scaling with exponent one, both the
attention and FFN residual updates are multiplied by $m_L^{-1}$, and the
hidden states entering the factorized readout are scaled by $m_W^{-1}$. For a hidden matrix, let $m_p$ be
its fan-in multiplier: $m_p=m_W$ except for the FFN down projection, where
$m_p=F/F_0$. This distinction accounts for rounding the FFN dimension to a
multiple of 256. Given base learning rate $\eta$, weight decay $\lambda$,
and AdamW stabilizer $\epsilon$, hidden matrices use
\begin{equation}
  \eta_p=\eta/m_p,\qquad
  \lambda_p=\lambda m_p,\qquad
  \epsilon_p=\epsilon/(m_p m_L).
  \label{eq:ar_hp_transfer_rules}
\end{equation}
Hidden normalization gains and biases retain base learning rate, receive
no weight decay, and use $\epsilon/(m_Wm_L)$. Input/output parameters use
base learning rate and $\epsilon/m_W$, with base weight decay on matrices
and no decay on one-dimensional parameters. The second text projection is
treated as a hidden matrix. There is no additional depth-dependent learning-rate
factor at exponent one. This experiment varies the base learning rate
under these fixed rules; it does not search over initialization scales.

\paragraph{Learning-rate grid and training recipe.}
We test the following 12 base learning rates for every architecture:
\begingroup
\scriptsize
\[
 \eta\in\{10^{-5},\,2\times10^{-5},\,3\times10^{-5},\,5\times10^{-5},\,
  7\times10^{-5},\,10^{-4},\,4\times10^{-4},\,6\times10^{-4},\,
  8\times10^{-4},\,10^{-3},\,2\times10^{-3},\,3\times10^{-3}\}.
\]
\endgroup
Each arm runs for 50,000 optimizer updates with global batch size 256
(32 images per rank on eight ranks), using the same in-house image--text
data recipe at the native-resolution $256^2$ tier. AdamW uses
$\beta_1=\beta_2=0.95$, base $\epsilon=10^{-8}$, base weight decay 0.01,
and gradient clipping at norm 1. The learning rate increases linearly
from $10^{-8}$ to the arm's base rate over 2,000 updates, then remains
constant, with the parameter-group multipliers applied throughout.
We use seed 42, text-condition dropout probability 0.1, BF16 model
computation.

\paragraph{Loss aggregation and recipe selection.}
Each arm trains for 50,000 updates. Each plotted point averages 21 raw
training CE measurements.
Repeated records at an optimizer step are resolved using the latest
timestamp before averaging. The shared reference architecture is plotted
in both panels. The statistic is training CE rather than held-out loss or decoded
image quality. Selecting $\eta=7\times10^{-5}$ gives the grid minimum at
all six depths and a maximum excess CE of 0.0048 nats per token across
all 11 architectures.

\begin{figure}[t]
  \centering
  \begin{minipage}{0.49\linewidth}
    \centering
    \includegraphics[width=\linewidth]{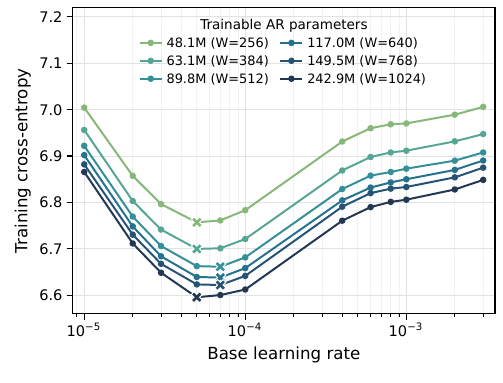}
    \small (a) Width transfer ($L=16$)
  \end{minipage}\hfill
  \begin{minipage}{0.49\linewidth}
    \centering
    \includegraphics[width=\linewidth]{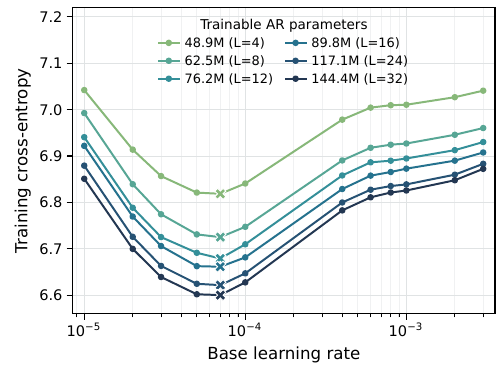}
    \small (b) Depth transfer ($W=512$)
  \end{minipage}
  \caption{\textbf{Pretraining learning-rate transfer across width and depth.}
  Each run trains for 50,000 updates. Each point averages
  21 raw training cross-entropy measurements; crosses mark the minimum among the 12 tested base
  learning rates.}
  \label{fig:ar_hp_transfer}
\end{figure}

\subsection{AR Scaling Protocol and Results}
\label{app:ar_protocol}

\paragraph{Training setup.}
We fit the 18 AR models in Tab.~\ref{tab:ar_scaling_models} using the
initialization and optimizer scaling rules in Appendix~\ref{app:ar_transfer}.
All models use a frozen IBQ tokenizer trained on Qwen2.5-VL-32B post-merger features
(131,072 codes, dimension 256, update-80,000 checkpoint), frozen FLAN-T5-XL
text features (maximum length 512), and images at the $256^2$ area tier.
Training uses AdamW with learning rate $7\times10^{-5}$,
$\beta_1=\beta_2=0.95$, $\epsilon=10^{-8}$, weight decay 0.01,
gradient clipping at norm 1, and batch size 1,024.
The learning rate warms up for 2,000 updates and remains constant thereafter.

\begin{table}[t]
 \centering
 \caption{\textbf{AR architectures used for the scaling fits.}
 Models are sorted by trainable parameters $N$; $W$ and $F$ denote hidden
 and FFN dimensions. Shared vocabulary weights are counted once;
 frozen encoders and tokenizers are excluded.}
 \label{tab:ar_scaling_models}
 \small
 \setlength{\tabcolsep}{7pt}
 \renewcommand{\arraystretch}{1.12}
 \begin{tabular*}{\linewidth}{@{\extracolsep{\fill}}rrrrrr@{}}
 \toprule
 $N$ (M) & Layers (depth) & Hidden dim. $W$ & Heads & Head dim. & FFN dim. $F$ \\
 \midrule
 55.71 & 6 & 512 & 4 & 128 & 1,536 \\
 71.30 & 7 & 640 & 5 & 128 & 1,792 \\
 99.96 & 9 & 768 & 6 & 128 & 2,048 \\
 137.73 & 10 & 896 & 7 & 128 & 2,560 \\
 191.53 & 12 & 1,024 & 8 & 128 & 2,816 \\
 260.95 & 14 & 1,152 & 9 & 128 & 3,072 \\
 323.07 & 14 & 1,280 & 10 & 128 & 3,328 \\
 343.39 & 15 & 1,280 & 10 & 128 & 3,584 \\
 401.57 & 15 & 1,408 & 11 & 128 & 3,840 \\
 425.72 & 16 & 1,408 & 11 & 128 & 3,840 \\
 493.02 & 16 & 1,536 & 12 & 128 & 4,096 \\
 549.65 & 18 & 1,536 & 12 & 128 & 4,096 \\
 688.28 & 19 & 1,664 & 13 & 128 & 4,608 \\
 860.44 & 21 & 1,792 & 14 & 128 & 4,864 \\
 1,059.83 & 23 & 1,920 & 15 & 128 & 5,120 \\
 1,276.36 & 24 & 2,048 & 16 & 128 & 5,632 \\
 1,764.95 & 27 & 2,304 & 18 & 128 & 6,144 \\
 2,425.92 & 30 & 2,560 & 20 & 128 & 6,912 \\
 \bottomrule
 \end{tabular*}
\end{table}

\paragraph{Loss and compute.}

Let $D_{\rm text}$ count text tokens and $D_{\rm tgt}$ count supervised
visual targets. We count all processed sequence positions:
\begin{equation}
 D_{\rm seq}=D_{\rm text}+2D_{\rm tgt},\qquad C=6ND_{\rm seq}.
 \label{eq:ar_compute_conventions}
\end{equation}
The factor of two accounts for one position-instruction token per visual code.

\paragraph{Training-curve envelope.}
Following \citet{hoffmann2022computeoptimal,ge2026chimera}, we interpolate
loss linearly in $\log C$ within each model's observed range and select
the minimum at 1,600 log-spaced budgets. The range starts at the first
crossing of the two smallest models and ends at the smaller of
$2\times10^{21}$ and the maximum observed compute.
Each winner defines $N_{\rm opt}$ and $D_{\rm opt}=C/(6N_{\rm opt})$.
We fit power laws by ordinary least squares in log--log space, retaining
boundary winners. Figures display one point per winning model, nearest
the geometric mean of its winning budgets.

\paragraph{IsoFLOP profiles.}
We interpolate loss at each budget and fit a quadratic in $\log N$.
A valid minimum requires positive curvature, at least five models, and
at least two model sizes on either side. The candidate budgets are
\begin{equation*}
 C/10^{19}\in\{0.3,0.5,0.75,1,1.5,2,3,5,7.5,10,15,20,30\}.
\end{equation*}
We regress $N_{\rm opt}$ and $D_{\rm opt}=C/(6N_{\rm opt})$ in log--log
space over the longest contiguous valid segment.

\paragraph{Parametric loss.}
Within the envelope range, we interpolate loss in $\log D$ at 24
geometrically spaced token counts per model after update 5,000.
For the resulting 432 equally weighted samples, we fit
$\widehat{\mathcal L}=E+An^{-\alpha}+Bd^{-\beta}$,
where $n=N/10^8$ and $d=D/10^9$, by minimizing
\begin{equation}
 \frac{1}{432}\sum_{i=1}^{432}H_{10^{-3}}
 (\log\widehat{\mathcal L}_i-\log\mathcal L_i),
 \label{eq:ar_huber_fit}
\end{equation}
where $H_\delta$ is the Huber loss. We select the lowest objective among
75 L-BFGS-B initializations with positive coefficients and exponents.
The optimal allocation has $a=\beta/(\alpha+\beta)$ and $b=1-a$.

\paragraph{Full-sequence results.}
The envelope covers $1.5535\times10^{18}$--$2\times10^{21}$ FLOPs;
IsoFLOP uses 12 budgets over $5\times10^{18}$--$3\times10^{20}$ FLOPs.
The parametric loss coefficients are
\begin{align*}
 E&=5.7238, & A&=0.3194, & B&=1.7284,\\
 \alpha&=0.4707, & \beta&=0.4386.
\end{align*}

Tab.~\ref{tab:ar_allocation_laws} gives the full-sequence allocation laws.

\begin{table}[htbp]
 \centering\small
 \caption{\textbf{Full-sequence AR allocation laws.}
 $N=N_{20}(C/10^{20})^a$ and $D=D_{20}(C/10^{20})^b$.}
 \label{tab:ar_allocation_laws}
 \setlength{\tabcolsep}{5pt}
 \begin{tabular*}{\linewidth}{@{\extracolsep{\fill}}lrrrr@{}}\toprule
 Method & $N_{20}$ (M) & $D_{20}$ (B) & $a$ & $b$ \\
 \midrule
 Envelope & 296.96 & 56.12 & 0.5268 & 0.4732 \\
 IsoFLOP & 270.58 & 61.60 & 0.4705 & 0.5295 \\
 Parametric & 199.05 & 83.73 & 0.4823 & 0.5177 \\
 \bottomrule\end{tabular*}
\end{table}

\subsection{Post-training Scaling Protocol and Results}
\label{app:post_scaling_results}

\paragraph{Training recipe.}
We initialize each model from its pretrained AR backbone and keep the AR
parameters, including attention projections, frozen. Only the diffusion
FFNs, AdaLN modules, and input/output adapters are newly initialized and
trained; the adapters include patch and timestep embeddings and the pixel
readout. We use CompleteP initialization with base scale 0.02.

The scaling family uses native images at the $256^2$ area tier,
14-pixel diffusion patches, frozen FLAN-T5-XL text features with maximum
length 512, and global batch size 512. We train with AdamW
($\beta_1=\beta_2=0.95$, $\epsilon=10^{-8}$, weight decay 0.01), clip gradients
at norm 1, and use base learning rate $8\times10^{-4}$ without warmup or
learning-rate decay. Parameter-group rates follow CompleteP scaling.
Parameter EMA is disabled.

We normalize images to $[-1,1]$, use Gaussian noise with standard deviation
2, and sample the clean-time coordinate from a logit-normal distribution
with location 0 and scale 1. The velocity conversion uses denominator
$\max(1-u,0.05)$, as in Appendix~\ref{app:tokenizer_controls}.
Text conditioning is dropped with probability 0.1; the AR visual
conditioning is retained.

\paragraph{Post-training learning-rate search.}
We search the eight base learning rates
$\{5\times10^{-5},10^{-4},4\times10^{-4},6\times10^{-4},
8\times10^{-4},10^{-3},2\times10^{-3},3\times10^{-3}\}$
at width/depth pairs $(512,6)$, $(768,9)$, $(1024,12)$, and $(1280,15)$.
For each architecture, the pretrained AR checkpoint and diffusion
initialization are fixed across learning rates. We compare all eight rates
for all four architectures at approximately 20,000 updates, averaging
41 raw loss measurements without smoothing. For the three smaller models,
the window spans updates 19,600--20,400. For the largest model, we use the
last 41 measurements of each run: updates 18,060--18,860 for the four
lower rates and 18,480--19,280 for the four higher rates.

The observed minima occur at $10^{-3}$, $6\times10^{-4}$,
$6\times10^{-4}$, and $8\times10^{-4}$ in order of increasing model size.
The shared $8\times10^{-4}$ rate used for scaling is within 0.0003
velocity-MSE of the observed minimum for all four models.

\begin{figure}[htbp]
\centering
\includegraphics[width=0.55\linewidth]{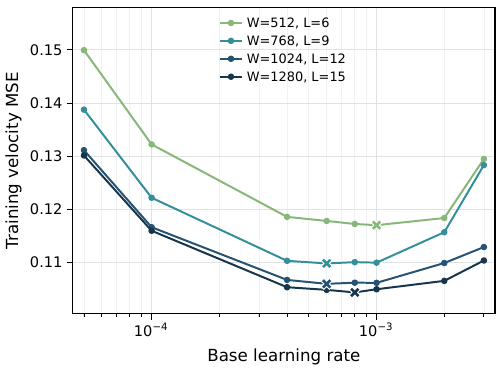}
\caption{\textbf{Post-training learning-rate search.} Mean training loss
at approximately 20,000 updates for four width/depth pairs. Each point
averages 41 measurements; the largest model uses its final available
windows. Crosses mark the minimum for each model.}
\label{fig:post_hp_transfer}
\end{figure}

\paragraph{Data and parameter accounting.}
We fit the 14 architectures in Tab.~\ref{tab:post_scaling_models} using
raw training velocity-MSE, without smoothing or a loss threshold.
The results use the fixed September 20 data snapshot. Each renderer is
trained on its corresponding frozen AR backbone. We first fit the AR
scaling law and, for each model size, select the AR checkpoint corresponding
to its predicted compute-optimal FLOPs budget for diffusion post-training.

\begin{table}[htbp]
\centering\small
\caption{Post-training model family, ordered by trainable diffusion parameter
count. Attention heads have dimension 128. The AR backbone and external
encoders remain frozen and are excluded from the parameter counts.}
\label{tab:post_scaling_models}
\begin{tabular}{rrrrr}\toprule
Parameters (M) & Layers & Width & Heads & FFN width\\\midrule
25.13 & 6 & 512 & 4 & 1536 \\
43.46 & 7 & 640 & 5 & 1792 \\
77.23 & 9 & 768 & 6 & 2048 \\
120.73 & 10 & 896 & 7 & 2560 \\
184.00 & 12 & 1024 & 8 & 2816 \\
265.85 & 14 & 1152 & 9 & 3072 \\
360.77 & 15 & 1280 & 10 & 3584 \\
457.95 & 16 & 1408 & 11 & 3840 \\
604.00 & 18 & 1536 & 12 & 4096 \\
763.60 & 19 & 1664 & 13 & 4608 \\
966.18 & 21 & 1792 & 14 & 4864 \\
1201.11 & 23 & 1920 & 15 & 5120 \\
1450.28 & 24 & 2048 & 16 & 5632 \\
2026.19 & 27 & 2304 & 18 & 6144 \\
\bottomrule
\end{tabular}
\end{table}

$N$ counts trainable diffusion parameters, including embeddings, modulation,
feed-forward blocks, and readout, but excludes the frozen AR path and external
encoders. Let $D_{\rm AR}$ count the AR codes consumed during post-training,
$D_{\rm tgt}$ the supervised diffusion patches, and $D_{\rm text}$ the text
positions. Each AR code has one position-instruction token. An AR grid of
$g_h\times g_w$ covers an image of $28g_h\times28g_w$ pixels, whereas the
renderer uses $14\times14$ patches. Thus each image contains four diffusion
patches per AR code, and the cumulative counts satisfy
\begin{equation}
 D_{\rm tgt}=4D_{\rm AR},\qquad
 D_{\rm seq}=D_{\rm text}+2D_{\rm AR}+D_{\rm tgt}
             =D_{\rm text}+1.5D_{\rm tgt}.
\end{equation}
We use $C=6ND_{\rm seq}$ as a normalized compute proxy rather than a
physical-operation count: exact FLOPs require separate accounting
for the frozen AR forward pass and the trainable renderer's forward/backward
passes. The source AR
checkpoint varies across models, so the fitted law is conditional on this
checkpoint family.

\paragraph{Parametric loss modelling.}
The fit covers $5\times10^{17}$--$3.0449\times10^{21}$ in full-sequence
compute, ending at the maximum observed budget. The training-curve envelope
uses the same interval. We fit
$\widehat L=E+A(N/10^8)^{-\alpha}+B(D/10^{10})^{-\beta}$ by minimizing squared
log-loss residuals, giving each model equal total weight. All eligible observations enter the fit. We select the lowest objective
among nine initializations, constraining $E\geq0$ and the amplitudes and
exponents to be positive. The fitted surface is
\begin{equation}
 \widehat L_{\rm seq}
 =0.0701+0.0222(N/10^8)^{-0.2531}
   +0.0126(D_{\rm seq}/10^{10})^{-0.5964}.
\end{equation}
The allocation exponents satisfy
$a=\beta/(\alpha+\beta)$ and $b=1-a$ under $C=6ND$.

\paragraph{Training-curve envelope.}
We interpolate loss linearly in $\log C$ within each model's observed range,
select the lowest loss at 500 log-spaced budgets, and regress $\log N_{\rm opt}$
and $\log D_{\rm opt}$ on $\log C$. Boundary winners are retained. All budgets enter the
regression, whereas figures display one representative per winning model,
nearest the geometric mean of its winning budgets.

\paragraph{IsoFLOP profiles.}
We evaluate 15 log-spaced budgets over the same compute interval. At each budget, a quadratic in $\log(N/10^8)$ is fitted to
interpolated raw loss. An accepted minimum requires at least three model
sizes, positive curvature, and both the observed minimum and fitted vertex
strictly inside the available size range. We regress all accepted vertices:
eight budgets over $6.03\times10^{18}$--$1.63\times10^{21}$.
No training curve is extrapolated to construct an envelope or IsoFLOP point.

\begin{table}[htbp]
\centering\small
\caption{Post-training allocation fits. With $x=C/10^{20}$,
$N_{\rm opt}=N_{20}x^a$ and $D_{\rm opt}=D_{20}x^b$, using full-sequence
token counts.}
\label{tab:post_scaling_allocations}
\begin{tabular}{lrrrr}\toprule
Method & $N_{20}$ (M) & $D_{20}$ (B) & $a$ & $b$\\\midrule
Parametric & 513.48 & 32.46 & 0.7021 & 0.2979\\
Envelope & 395.28 & 42.16 & 0.5492 & 0.4508\\
IsoFLOP & 377.99 & 44.09 & 0.4538 & 0.5462\\
\bottomrule\end{tabular}
\end{table}

\subsection{Pixel-Diffusion Hyperparameter Transfer}
\label{app:pdm_hp}

\begin{figure}[t]
\centering
\begin{minipage}{0.49\linewidth}
  \centering
  \includegraphics[width=\linewidth]{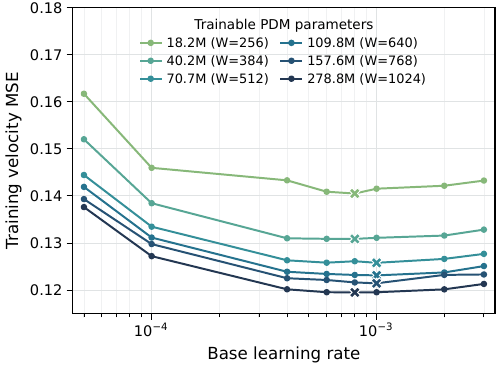}
  \small (a) Width transfer ($L=16$)
\end{minipage}\hfill
\begin{minipage}{0.49\linewidth}
  \centering
  \includegraphics[width=\linewidth]{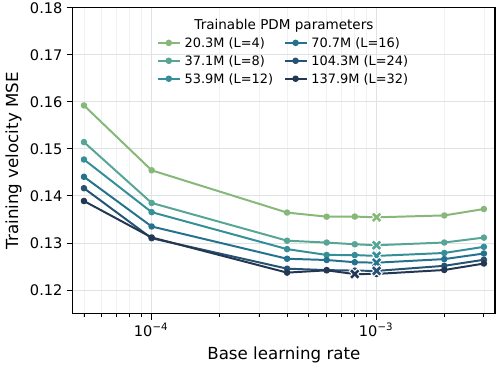}
  \small (b) Depth transfer ($W=512$)
\end{minipage}
\caption{\textbf{Pixel-diffusion learning-rate transfer.} Original width and
 depth search results after 50,000 training updates. Points average the
 final 10,000 updates; crosses mark the best tested learning rate.}
\label{fig:pdm_hp_transfer}
\end{figure}

We independently tune the PDM baseline trained from scratch using
CompleteP learning-rate transfer. We vary width over
$\{256,384,512,640,768,1024\}$ at 16 layers and depth over
$\{4,8,12,16,24,32\}$ at width 512. Each configuration is trained for
50,000 updates. The eight base learning rates are
\[
\{5\times10^{-5},10^{-4},4\times10^{-4},6\times10^{-4},
8\times10^{-4},10^{-3},2\times10^{-3},3\times10^{-3}\}.
\]
We retain the original search statistic: mean training velocity-MSE over
the final 10,000 updates. Each point averages 500 logged measurements,
without smoothing. Fig.~\ref{fig:pdm_hp_transfer} shows the complete
width and depth searches. Every curve reaches its grid minimum at either
$8\times10^{-4}$ or $10^{-3}$, supporting learning-rate transfer across
model sizes. Tab.~\ref{tab:pdm_hp_minima} lists these minima.

The original search uses 16-pixel patches and FFN width $4W$, where $W$
is model width. Additional checks at five anchor architectures use
FFN width rounded from $8W/3$ and learning rates
$\{6\times10^{-4},8\times10^{-4},10^{-3},1.3\times10^{-3}\}$.
Their minima also lie at $8\times10^{-4}$ or $10^{-3}$.
The scaling family uses this rounded FFN configuration with 14-pixel
patches and base learning rate $8\times10^{-4}$. Training uses global
batch size 512, AdamW with $\beta_1=\beta_2=0.95$, $\epsilon=10^{-8}$,
weight decay 0.01, and 2,000 warmup updates followed by a constant rate.
This search concerns PDMs trained from scratch. The independent post-training
search is described in Appendix~\ref{app:post_scaling_results}.

\begin{table}[htbp]
\centering\small
\caption{Grid minima in the original PDM hyperparameter search. The shared
width-512, depth-16 architecture is evaluated in both sweeps.}
\label{tab:pdm_hp_minima}
\begin{tabular}{lrrrr}\toprule
Sweep & Width & Layers & Best base LR & Training loss\\\midrule
Width & 256 & 16 & $8\times10^{-4}$ & 0.1405 \\
Width & 384 & 16 & $8\times10^{-4}$ & 0.1309 \\
Width & 512 & 16 & $10^{-3}$ & 0.1258 \\
Width & 640 & 16 & $10^{-3}$ & 0.1231 \\
Width & 768 & 16 & $10^{-3}$ & 0.1214 \\
Width & 1024 & 16 & $8\times10^{-4}$ & 0.1195 \\
Depth & 512 & 4 & $10^{-3}$ & 0.1355 \\
Depth & 512 & 8 & $10^{-3}$ & 0.1295 \\
Depth & 512 & 12 & $10^{-3}$ & 0.1273 \\
Depth & 512 & 16 & $10^{-3}$ & 0.1258 \\
Depth & 512 & 24 & $10^{-3}$ & 0.1241 \\
Depth & 512 & 32 & $8\times10^{-4}$ & 0.1234 \\
\bottomrule\end{tabular}
\end{table}

\subsection{Pixel-Diffusion Scaling from Scratch}
\label{app:pdm_scratch}

\paragraph{Models and compute.}
We train eight PDMs from scratch using the recipe
in Appendix~\ref{app:pdm_hp}. Noise scale, timestep distribution,
text-conditioning dropout, and loss averaging are identical to diffusion
post-training (Appendix~\ref{app:post_scaling_results}).
Tab.~\ref{tab:pdm_scaling_models} gives their
architectures and trainable parameter counts. The models contain no AR
backbone. Here $N$ includes the denoiser's attention, FFN, modulation,
text projection, patch embedding, and image readout, excluding the frozen
text encoder. $D_{\rm seq}$ counts the actual packed text and diffusion
positions, including the effect of text-condition dropout, and
$C=6ND_{\rm seq}$.

\begin{table}[htbp]
\centering\small
\caption{Pure PDM scaling architectures, ordered by trainable parameter
count. All attention heads have dimension 128.}
\label{tab:pdm_scaling_models}
\begin{tabular}{rrrrr}\toprule
Parameters (M) & Layers & Width & Heads & FFN width\\\midrule
23.83 & 6 & 512 & 4 & 1536 \\
69.97 & 9 & 768 & 6 & 2048 \\
164.08 & 12 & 1024 & 8 & 2816 \\
319.16 & 15 & 1280 & 10 & 3584 \\
529.29 & 18 & 1536 & 12 & 4096 \\
844.62 & 21 & 1792 & 14 & 4864 \\
1265.74 & 24 & 2048 & 16 & 5632 \\
1760.20 & 27 & 2304 & 18 & 6144 \\
\bottomrule\end{tabular}
\end{table}

\paragraph{Fitting protocol.}
We use raw training velocity-MSE without EMA, binning, or a loss threshold.
The envelope and parametric fits cover
$5\times10^{17}\le C\le2.2174\times10^{21}$.
We interpolate each observed loss curve in $\log C$ and select the minimum
at 500 log-spaced budgets. All envelope winners, including boundary winners,
enter the log--log allocation regression; figures display one representative
point per winning model.

For IsoFLOP, we evaluate 15 log-spaced candidate budgets and fit a quadratic
in $\log(N/10^8)$ at each budget. A valid minimum requires at least three
model sizes, positive curvature, and both the observed minimum and fitted
vertex inside the available size range. Nine budgets are accepted over
$1.6594\times10^{18}$--$6.6813\times10^{20}$ FLOPs. No training curve is
extrapolated to construct an envelope or IsoFLOP point.

\begin{figure}[t]
\centering
\includegraphics[width=\linewidth]{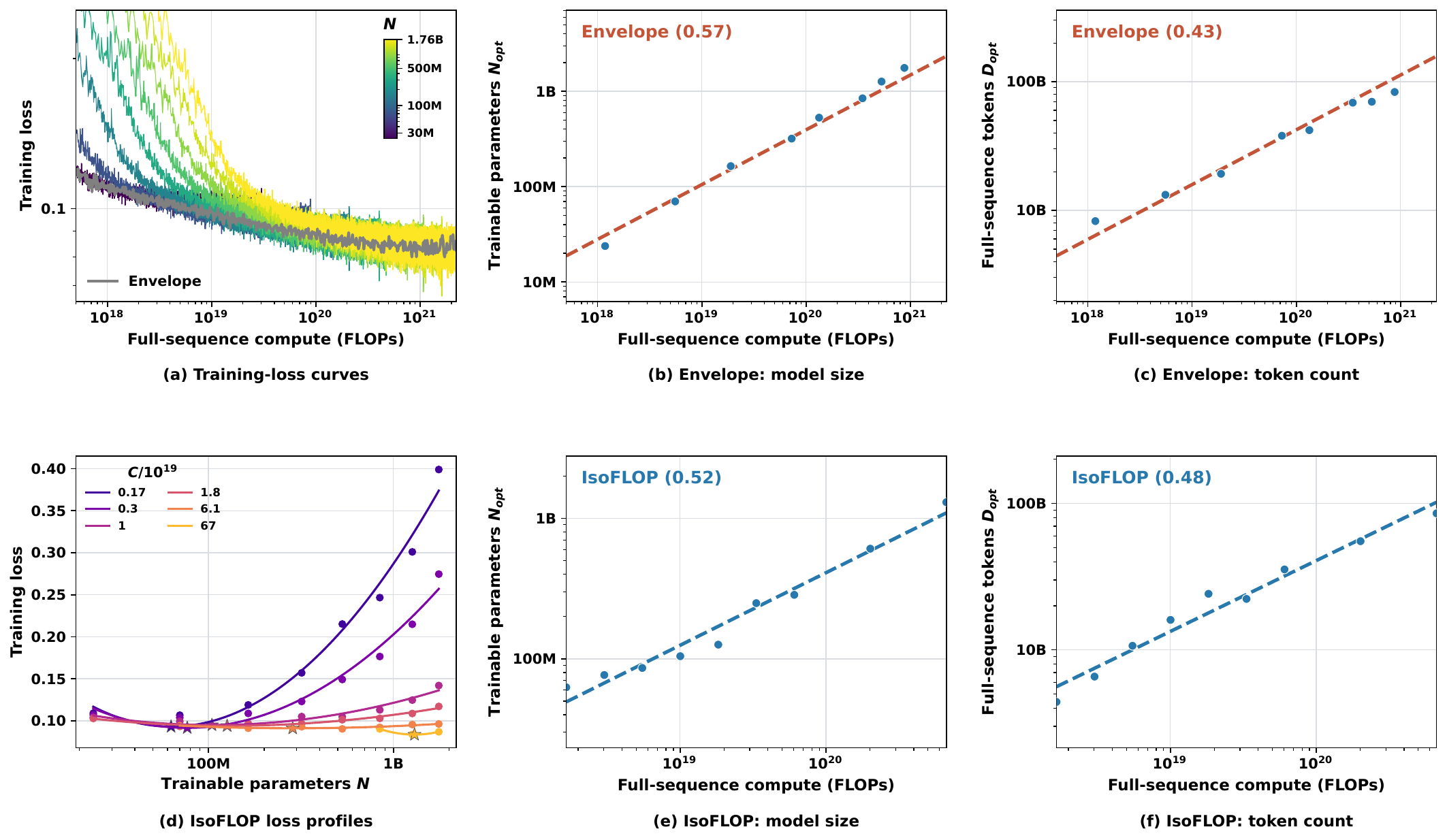}
\caption{\textbf{Pixel-diffusion scaling from scratch.} Full-sequence
compute accounting for eight model sizes. Top: training losses and envelope
allocations. Bottom: six representative IsoFLOP profiles and allocation
fits using all nine accepted minima. Stars mark the quadratic minima.}
\vspace{-5mm}
\label{fig:pdm_scratch_scaling}
\end{figure}

For the parametric estimator, we fit all eligible observations by minimizing
squared log-loss residuals, giving each model equal total weight. We retain
the best of nine initializations, with $E\ge0$ and positive amplitudes and
exponents. The resulting loss law is
\begin{equation}
\widehat{\mathcal L}_{\rm scratch}
=0.0788+0.0133\left(\frac{N}{10^8}\right)^{-0.3450}
+0.0079\left(\frac{D_{\rm seq}}{10^{10}}\right)^{-0.8627}.
\end{equation}
Its fitted asymptotic loss exceeds the post-training value
$E_{\rm post}=0.0701$. Tab.~\ref{tab:pdm_scaling_allocations} gives the
three compute-optimal allocation fits, and
Fig.~\ref{fig:pdm_scratch_scaling} shows the envelope and IsoFLOP results.

\begin{table}[htbp]
\centering\small
\caption{Full-sequence PDM allocation laws. For $x=C/10^{20}$,
$N_{\rm opt}=N_{20}x^a$ and $D_{\rm opt}=D_{20}x^b$.}
\label{tab:pdm_scaling_allocations}
\begin{tabular}{lrrrr}\toprule
Method & $N_{20}$ (M) & $D_{20}$ (B) & $a$ & $b$\\\midrule
Envelope & 394.37 & 42.26 & 0.5744 & 0.4256\\
IsoFLOP & 409.73 & 40.68 & 0.5161 & 0.4839\\
Parametric & 535.04 & 31.15 & 0.7143 & 0.2857\\
\bottomrule\end{tabular}
\end{table}

\subsection{Compute Accounting and Stage Allocation}
\label{app:post_protocol}

We use $C=6ND$ throughout, with $N$ counting the trainable parameters of
that stage and $D$ counting processed sequence positions. For diffusion
post-training, $N$ excludes the frozen AR backbone; for diffusion from
scratch, it includes the entire denoising model. Both exclude external
frozen encoders. Pure PDM processes text and diffusion patches, whereas
post-training also processes AR codes and position-instruction tokens.
We therefore use each model's actual packed-token count rather than
assigning the same token count to the two architectures.

\paragraph{Induced budget allocation.}
We convert the two envelope laws into a budget split. Inverting
$N=N_{20}(C/10^{20})^{a}$ gives the compute at which a model of size $N$ is the
envelope winner, $C^\star(N)=10^{20}(N/N_{20})^{1/a}$. For each jointly scaled
model we evaluate $C_{\mathrm{pre}}^\star(N_{\mathrm{AR}})$ from
Appendix~\ref{app:ar_protocol} and $C_{\mathrm{post}}^\star(N_{\mathrm{diff}})$
from Appendix~\ref{app:post_scaling_results} using full-sequence counts,
then set $r=C_{\mathrm{post}}^\star/(C_{\mathrm{pre}}^\star+C_{\mathrm{post}}^\star)$.
Tab.~\ref{tab:compute_allocation}
lists the nine models with $N_{\mathrm{AR}}\ge261$M; the smaller models give
noisy envelope winners and unstable splits. Summed over these models the
split is $C_{\mathrm{pre}}:C_{\mathrm{post}}=1.50{:}1$ ($r=0.40$).
This allocation uses the normalized $6ND$ proxy, which excludes the frozen-AR
forward pass in post-training, rather than measured hardware FLOPs.
AR cross-entropy is never combined numerically with diffusion loss;
each stage is optimized under its own metric.

\begin{table}[htbp]
  \centering\small
  \caption{\textbf{Induced pretraining--post-training budget allocation at the envelope
  compute-optimal.} For each jointly scaled model, each stage is placed at its
  own envelope optimum and $r=C_{\mathrm{post}}/(C_{\mathrm{pre}}+C_{\mathrm{post}})$.
  $N$ excludes frozen modules; $C$ uses the full-sequence $6ND$ proxy.
  Models below $261$M are omitted as noisy.}
  \label{tab:compute_allocation}
  \setlength{\tabcolsep}{4pt}
  \begin{tabular*}{\linewidth}{@{\extracolsep{\fill}}rrrrr@{}}\toprule
  $N_{\mathrm{AR}}$(M) & $N_{\mathrm{diff}}$(M) & $C_{\mathrm{pre}}$ & $C_{\mathrm{post}}$ & $r$\\
  \midrule
  260.95 & 265.85 & $7.82\times10^{19}$ & $4.86\times10^{19}$ & 0.383\\
  343.39 & 360.77 & $1.32\times10^{20}$ & $8.47\times10^{19}$ & 0.391\\
  425.72 & 457.95 & $1.98\times10^{20}$ & $1.31\times10^{20}$ & 0.398\\
  549.65 & 604.00 & $3.22\times10^{20}$ & $2.16\times10^{20}$ & 0.402\\
  688.28 & 763.60 & $4.93\times10^{20}$ & $3.32\times10^{20}$ & 0.402\\
  860.44 & 966.18 & $7.53\times10^{20}$ & $5.09\times10^{20}$ & 0.403\\
  1059.83 & 1201.11 & $1.12\times10^{21}$ & $7.57\times10^{20}$ & 0.403\\
  1276.36 & 1450.28 & $1.59\times10^{21}$ & $1.07\times10^{21}$ & 0.401\\
  1764.95 & 2026.19 & $2.95\times10^{21}$ & $1.96\times10^{21}$ & 0.400\\
  \midrule
  \multicolumn{2}{r}{Total} & $7.63\times10^{21}$ & $5.11\times10^{21}$ & 0.401\\
  \bottomrule\end{tabular*}
\end{table}

\subsection{Final Model Recipe and Evaluation}
\label{app:evaluation_protocol}
\label{app:hero_protocol}

\paragraph{Final-model training.}
The large-scale AR backbone has width 2304 and 27 layers. It uses frozen
Qwen3-4B text features from layers 9, 18, and 27, concatenated to 7680 dimensions;
feature extraction stops at layer 27 and uses BF16. AR pretraining uses a
global batch size of 2048 and a caption mixture of 70\% flattened long
captions and 30\% canonical structured captions.
The 150k-update AR curriculum uses 50k updates at the 256-resolution tier,
ramps toward a 50/25/25 mixture of the 256/512/1024 tiers over the next 50k,
and holds that mixture for the last 50k. These proportions count examples
within the global batch. This final-model recipe is separate from the
controlled comparisons in Tab.~\ref{tab:tokenizer_results}.
We then freeze the AR backbone and train the diffusion modules for 100,000
updates with global batch size 512. Diffusion uses four times as many
visual target tokens per image as AR pretraining, so we reduce its image
batch size by a factor of four. This balances visual-target throughput
per update at a matched resolution. Guided by the prescribed $3{:}2$
stage allocation, we use a matching ratio of pretraining to post-training
updates.
Post-training follows the diffusion resolution curriculum, preserving its
resolution proportions while reducing each resolution-specific batch size
proportionally to give global batch size 512. We use AdamW with weight decay
0.01 and base learning rate $8\times10^{-4}$, selected by the post-training
learning-rate search in Appendix~\ref{app:post_scaling_results}.

\paragraph{Training cost.}
The total training budget is approximately 6,600 H100 GPU-hours. Following
the prescribed $3{:}2$ allocation, the nominal stage budgets are approximately
3,960 GPU-hours for AR pretraining and 2,640 GPU-hours for diffusion
post-training. These are budget allocations rather than separate measurements
of stage runtime.

\paragraph{Evaluation sampling.}
We use an exponential moving average (EMA) of the trainable diffusion
parameters, with decay 0.9999 starting at update 20,000; the frozen AR
parameters remain unchanged.
We set AR classifier-free guidance (CFG) to 3 and use a 50-step Euler
sampler for diffusion, with AR-conditioning CFG 3 and text-conditioning
CFG 4. GenEval uses 553 prompts with one image per prompt and seed 42;
DPG-Bench uses 1,065 prompts with four images per prompt and seeds
42--45.
We use LLM-assisted prompt rewriting and select prompt variants using evaluation scores under fixed seeds.

\begin{figure*}[!t]
  \centering
  \includegraphics[width=\textwidth]{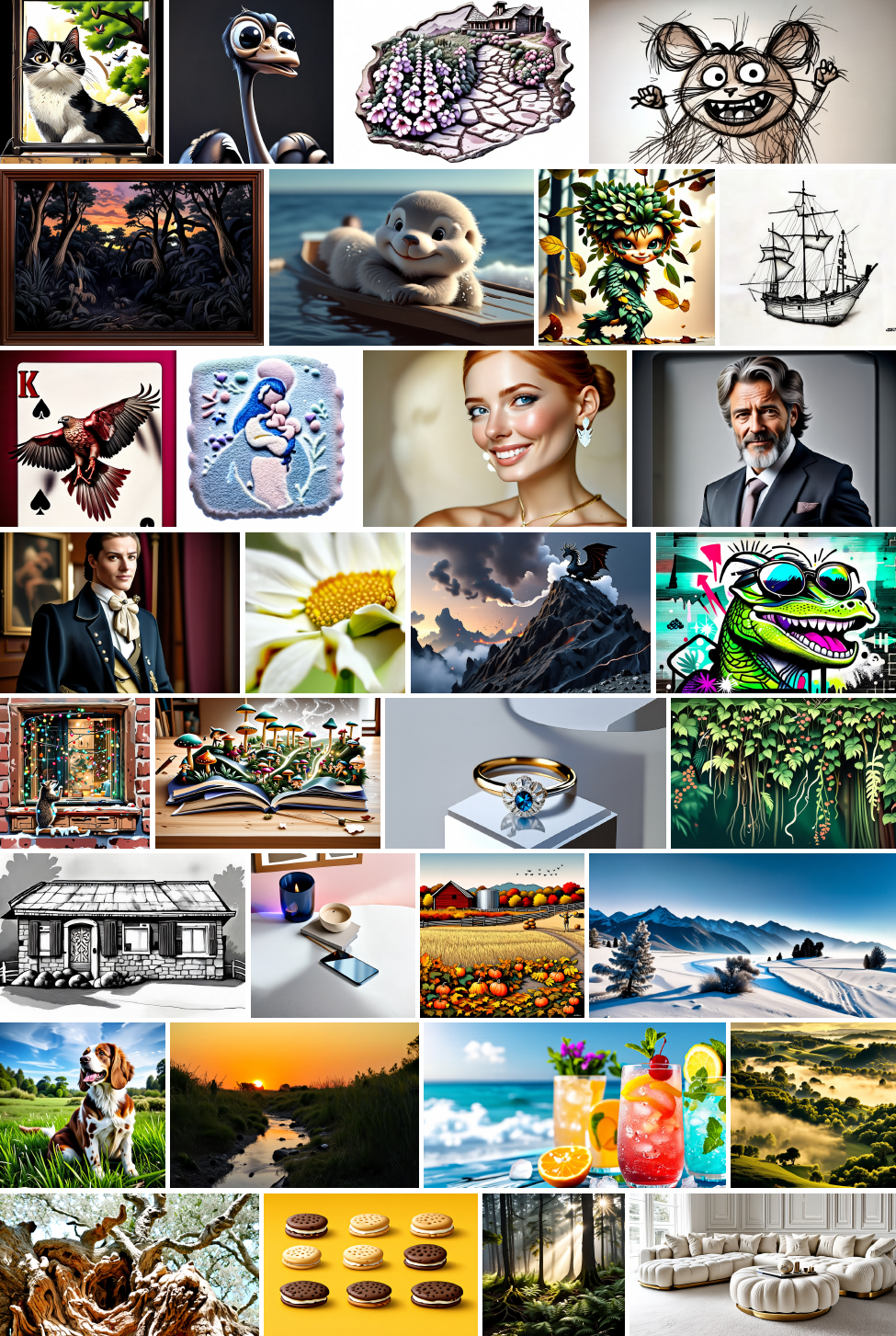}
  \caption{\textbf{Selected generation samples from our final text-to-image model.}
  }
  \label{fig:hero_samples}
\end{figure*}

\paragraph{Qualitative examples.}
Fig.~\ref{fig:hero_samples} presents 32 generations result from
the final text-to-image model.

\subsection{Discussion: Semantic inference and rendering in frontier models.}
Our results suggest a broader interpretation of the separation between knowledge
acquisition and rendering. We suspect that frontier visual generators~\citep{openai2026gptimage2,ideogram4,reve2026} increasingly perform substantial
semantic inference before committing to pixel generation. An autoregressive
process can expand a user request into a richer specification through text or
visual tokens, structured representations, retrieved information, or agentic
interaction. Prompt rewriting and long textual rollouts are simple instances of
this principle~\citep{chen2026scaling}: additional context progressively
specifies the desired image and reduces the ambiguity left to the renderer.
Under this view, the role of autoregression is not necessarily to produce an
image that can already be decoded, but to resolve \emph{what} should be
rendered; diffusion can then focus on realizing this specification in pixels.

\textbf{Semantic-LM admits a similar interpretation.}
Its discrete visual tokens are trained through language supervision and need not
reconstruct or directly generate an image. They can instead be viewed as a
compressed semantic state between inference and rendering. In this sense, a
semantic visual rollout and a long textual rollout may serve related purposes:
both enrich the renderer's conditioning context, but visual tokens may encode
the relevant information more compactly and in a spatially grounded form. This
perspective also raises a question not resolved by our current experiments.
Semantic-Recon retains reconstruction-oriented visual information, whereas
Semantic-LM more directly aligns its representation with language semantics.
Without explicit textual rollouts, our results do not determine which interface
is preferable when a generator already has access to rich language or agentic
context. We suspect that Semantic-LM may be particularly compatible with such
inference, while Semantic-Recon may preserve complementary visual information.
Understanding how textual, visual, structured, and agentic representations
should interact---and how much uncertainty each should resolve before
rendering---is an important direction for future visual generation systems.

\clearpage
\twocolumn
% The appendix ends with a two-column float page; keep its gallery at the top.
\makeatletter
\setlength{\@dblfptop}{0pt}
\makeatother

\end{document}